\documentclass{article}

\usepackage[preprint]{corl_2024} % Use this for the initial submission.

\usepackage{times}

\usepackage[numbers]{natbib}
\usepackage{multicol}
\usepackage{graphicx}
\usepackage{wrapfig}
\usepackage{amssymb}
\usepackage[normalem]{ulem}
\usepackage{xspace}
\usepackage[dvipsnames]{xcolor}
\usepackage{amsmath}
\usepackage[toc,page]{appendix}

\newcommand{\method}{\textit{Bellman-Guided Retrials}\xspace}

\usepackage{xcolor}
\definecolor{myblue}{rgb}{0.153, 0.66, 0.88}
\definecolor{mygreen}{rgb}{0.12, 0.54, 0.30}
\definecolor{myred}{rgb}{0.78, 0.33, 0.37}
\definecolor{myorange}{rgb}{0.98, 0.65, 0.15}
\hypersetup{
    colorlinks=true,
    linkcolor=myblue,
    filecolor=magenta,      
    urlcolor=myblue,
    citecolor=myblue,
}

\usepackage[ruled,vlined,commentsnumbered,titlenotnumbered]{algorithm2e}
\usepackage[noend]{algorithmic}
\usepackage{booktabs}

\begin{document}

% paper title
% \title{Rapid Robotic Trial and Error With Time-To-Success Models}

\title{To Err is Robotic: Rapid Value-Based Trial-and-Error during Deployment}

% You will get a Paper-ID when submitting a pdf file to the conference system
% \author{Author Names Omitted for Anonymous Review. Paper-ID 159}
% \author{Maximilian Du, Alexander Khazatsky, Tobias Gerstenberg, Chelsea Finn}

\author{
  Maximilian Du$^1$ \quad Alexander Khazatsky$^1$ \quad Tobias Gerstenberg$^2$ \quad Chelsea Finn$^1$\\
  $^1$Department of Computer Science \quad $^2$Department of Psychology \\
  Stanford University\\
  \texttt{\{maxjdu,alexkhaz,gerstenberg,cbfinn\}@stanford.edu} \\
}

% Change the focus to instilling "retrying" behaviors into robots
\newcommand{\ROne}{\textbf{\textcolor{blue}{n4x6}}}
\newcommand{\RTwo}{\textbf{\textcolor{olive}{vAmW}}}
\newcommand{\RThree}{\textbf{\textcolor{red}{AdQ3}}}

% \subsubsection{Some implementation details, such as policy parameterization and environmental reset mechanisms, are underexplored (Also raised by \ROne, \RTwo, \RThree) }

% \method provides a general framework of error detection and retrying. The components (base policy, value function, reset mechanisms, skewing) are fully modular, which means that we can easily swap in better-performing model for any of these components into our overarching framework. This is especially true for environment resets: we showed that even a simple pullback of the robot arm can improve performance by allowing the robot to retry. An adaptive resetting policy will likely improve that performance. Likewise, a more adaptive skewing method may be used in more dynamic environments. 

% During our experimentation, we chose a diffusion policy because it allows us to quickly sample multiple trajectories. We experimented with Transformer, LSTM, and simple feed-forward architectures, but they all performed worse than the diffusion policy. As we are most interested in relative performance changes between the base policy and our method, we picked the most competitive architecture as the base policy. 

\maketitle

\begin{abstract}
    When faced with a novel scenario, it can be hard to succeed on the first attempt. In these challenging situations, it is important to know how to \textit{retry} quickly and meaningfully. Retrying behavior can emerge naturally in robots trained on diverse data, but such robot policies will typically only exhibit undirected retrying behavior and may not terminate a suboptimal approach before an unrecoverable mistake. 
    We can improve these robot policies by instilling an explicit ability to try, evaluate, and retry a diverse range of strategies. 
    We introduce \method, an algorithm that works on top of a base robot policy by monitoring the robot's progress, detecting when a change of plan is needed, and adapting the executed strategy until the robot succeeds. 
    We start with a base policy trained on expert demonstrations of a variety of scenarios. 
    Then, using the same expert demonstrations, we train a value function to estimate task completion. During test time, we use the value function to compare our expected rate of progress to our achieved rate of progress. If our current strategy fails to make progress at a reasonable rate, we recover the robot and sample a new strategy from the base policy while skewing it away from behaviors that have recently failed. We evaluate our method on simulated and real-world environments that contain a diverse suite of scenarios. We find that \method increases the average absolute success rates of base policies by more than 20\% in simulation and 50\% in real-world experiments, demonstrating a promising framework for instilling existing trained policies with explicit trial and error capabilities. Refer to this site for evaluation videos: \href{https://sites.google.com/view/to-err-robotic/home}{https://sites.google.com/view/to-err-robotic/home} 
\end{abstract}


% \section*{TODO}
% \begin{itemize}
%     \item Better keywords
% \end{itemize}

\section{Introduction}

% On an icy January evening, snowboarder Lanny trudges to his motel room in the heart of the Adirondack Mountains. As he hastily knocks the snow from the old, rusty lock and reaches for his keys, let us consider how he tries to unlock this new door on his first try. He aligns the key to the keyhole, but it only goes in halfway before catching on something. He pulls the key out and tries again, this time wiggling the key up and down to break up the ice and rust in the keyhole. Slowly, the key squeezes into place. Lanny twists the key to the right, but nothing turns. Immediately, he twists it to the left, but still nothing happens. He twists right again, and this time, he feels the mechanism budge slightly. He gives the lock a much harder twist, and the deadbolt finally retracts. 

% Lanny unlocked his motel room in a matter of seconds, perhaps giving the illusion of simplicity. In fact, Lanny's scenario posed a difficult challenge. As he approached the new door, Lanny did not immediately know what would work. Instead, he tried out several different strategies for inserting a key and unlocking the door. He also had to monitor his own progress and switch strategies when the current strategy was making little progress. 

Real robots see many novel scenarios as they operate in the human world, which highlights the importance of rapid adaptation---and particularly, the ability to recover and retry after an initial mistake. Recent works have demonstrated that robots can learn sophisticated real-world strategies \cite{zhaoAloha, grannenStabilizeActLearning2023, xieDeepImitationLearning2020, smithDualArmManipulation2012}. 
% However, when faced with a novel situation, the robot may need to interact and even make mistakes to discover the correct strategy. 
Many such robot skills have been learned by imitating demonstrations collected by an expert \cite{zhaoAloha, schaalImitationLearningRoute1999}. From these demonstrations, the robot can implicitly learn how to recover and retry from a mistake. However, this ability depends on sufficient data diversity, and novel situations can pose a significant challenge. If the robot does not judiciously terminate an unproductive strategy, it may drift into an even more unfamiliar situation, which will further impair the performance of its policy \cite{rossReductionImitationLearning2011, tuSampleComplexityStability2022}, including any learned ability to recover and retry. Even if the robot is robust to failure states in this novel scenario, there is no guarantee that the robot will try strategies systematically.

% \begin{wrapfigure}{r}{0.45\textwidth}
% \begin{figure}[t]
%     \centering %centers everything 
%     \includegraphics[width = 0.44\textwidth]{figures/pullfig.pdf}
%     \caption{\small \textbf{Detecting Failures and Encouraging Diverse Attempts.} Leveraging a value function learned from expert demonstrations, our method helps a pretrained robot policy succeed in a novel scenario. In the example above, the pretrained policy is faced with a new object that it grasps improperly (1). Our method detects the failure and recovers the robot, allowing it to try again and modify its behavior to avoid the same mistake. The second attempt also leads to a failure (2) that we detect and recover from. The third attempt (3) is successful. With our method, the robot makes multiple, diverse attempts--- a capability that is important when faced with a new scenario.} 
%     \vspace{-0.5cm}
%     \label{fig:pullfig} 
% \end{figure}
% \end{wrapfigure}

Recognizing the shortcomings of learning retrying behavior implicitly from expert demonstrations, we propose \method, a method that endows expert-trained robot policies with \textit{explicit and systematic} strategy retrying behavior, allowing them to adapt quickly to novel scenarios. We design \method around a key insight: while attempting a novel situation, a robot needs to monitor its progress. If it is progressing slower than expected, then it should try something different. By keeping track of previous failures, we can avoid retrying these suboptimal strategies. 

We design \method to build on top of an existing base robot policy, which means that our method requires no external reward signals, expert interventions, or additional data outside of the pre-collected set of expert demonstrations used to train the base robot policy. We leverage the demonstrations to model good expert progress \cite{mazoureAcceleratingExplorationRepresentation2023a}. During test time, we can compare the robot's progress with this expert progress to evaluate how optimal the current strategy is.

Concretely, we train a value function $V_\phi$ to estimate the time it should take the expert demonstrator to solve the scenario from a given state. During task execution, we monitor how consistent the value function's predictions are. If the value function significantly overestimates the robot's true performance, then the current strategy is likely suboptimal. By constantly monitoring the robot's progress, we can stop the robot's execution of a bad strategy before it reaches an unrecoverable state. 
% Because we are checking for the self-consistency of $V_\phi$, the strategy evaluation can still detect mistakes that bring the robot to a previously unseen state. 

\begin{figure*}[t]
    \centering %centers everything 
    \includegraphics[width = 0.9\textwidth]{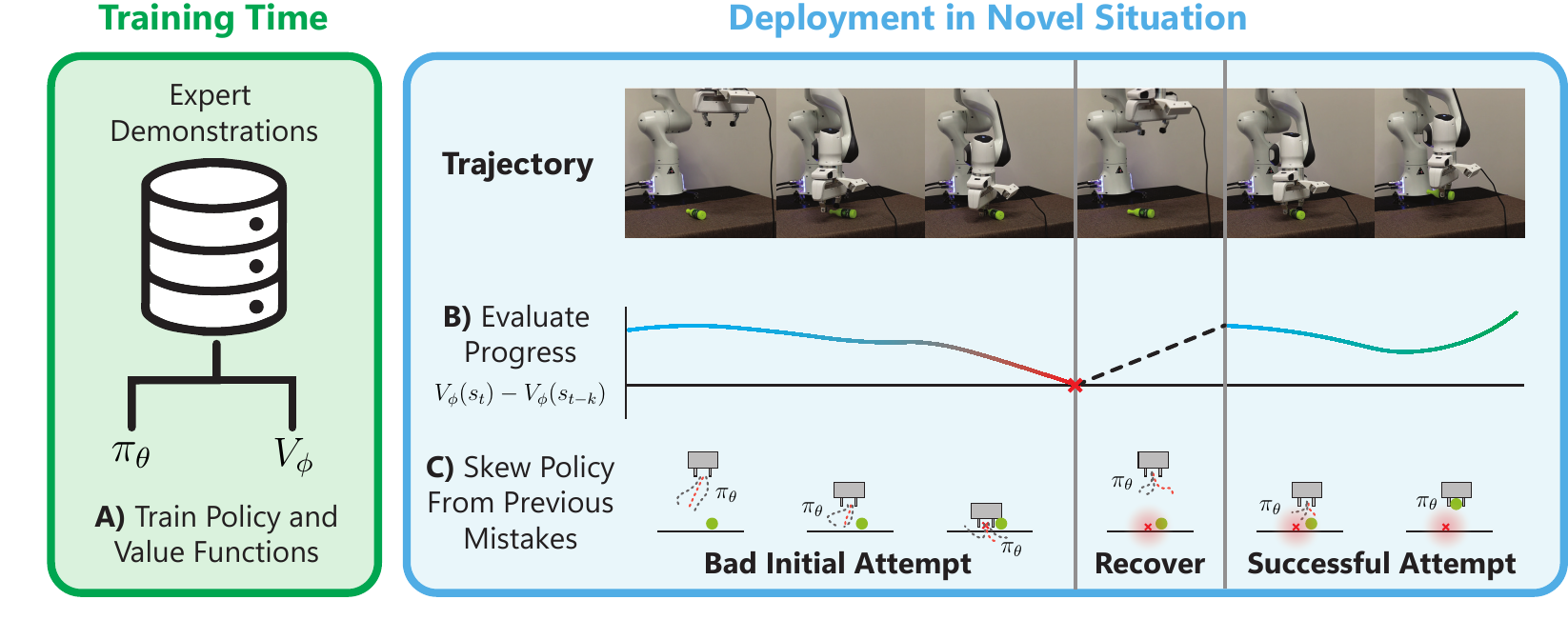}
    \vspace{-0.2cm}
    \caption{\small \textbf{The \method  Method. } \textbf{(A)} Using expert demonstrations, we train a policy and value function. While solving a novel situation, we evaluate progress by looking at the behavior of the trained value function \textbf{(B)}. If we detect suboptimality, we recover the robot and perform \textbf{(C)}, which modifies the sampling of the pretrained policy to avoid past mistake states.} 
        \vspace{-0.5cm}
    \label{fig:methodsfig} 
\end{figure*}

When we detect suboptimality, we execute a recovery policy. In many cases, this policy can be as simple as retracting a robot arm. Then, we sample another strategy from the base policy. We record states associated with suboptimal attempts and skew the sampling process to avoid similar states. The skew modification acts on top of the base policy's existing sampling process by sampling multiple action proposals and picking the proposal that is least likely to yield a suboptimal state. This skewing method is non-parametric and can work with very little interaction data. 

The main contribution of this work is a general framework for enabling explicit trial-and-error capabilities during deployment. Critically, by avoiding fine-tuning and using a separately trained value function, our approach is a drop-in replacement for the base policy.  
We test our method on two simulated grasping environments with a large collection of realistic objects. We perturb the test-time scenario by introducing novel objects or obstacles. We also test our method on a real robot in a similar grasping task, as well as a longer horizon door-opening task. Our approach boosts success rates by more than 20\% in our simulation evaluations and more than 50\% on a real robot setup.

% . The lack of parametric updates to the expert-trained policy also means that robots equipped with our method can successfully find a working strategy within just a few trials.

% In both simulation and the real world, we evaluate our framework across pick and place, door opening and closing, and XYZ. We find that on average, our framework increases the success rate of the underlying policy by N\% on robot hardware with under M seconds of interaction per episode, demonstrating exciting potential for robots to adapt rapidly in the real world.
%%CF.1.10: novel environment is a little vague/imprecise and open for misinterpretation -- usually it's a novel object right?
%Our method is entirely self-supervised,
%%CF.1.10: well, it uses demos... this claim is probably needlessly controversial. Or maybe it's just imprecise. It would be good to convey that *adaptation* doesn't require any human supervision.

\section{Related Work}

%%CF.1.10: I don't think you need reset-free RL (though you do need reset-free adaptation, like single life RL)
%%CF.1.10: I think the imitation learning section can be fairly brief. You may not even need it. We use IL as a subroutine, but it's not the focus of the paper.

% Shorted, add value function work, clean up generally
% The fact that our method involves finding a problem point is similar rto dagger and hdagger, but as apposed to these methods we dont require anything beyond the original data. 

\noindent \textbf{Learning From Demonstrations:} 
In our framework, we assume access to a dataset of expert robot task demonstrations. With this dataset, a common approach is to train a control policy to imitate the demonstrated expert behavior \cite{alvinn, argallSurveyRobotLearning2009, billardRobotProgrammingDemonstration2008,schaalComputationalApproachesMotor2003, schaalImitationLearningRoute1999}.
%%CF.1.31: should include some older cites like ALVINN, maybe some older Stefan Schaal stuff, a survey (maybe the Brenna Argal one) [resolved]
Extensive work has explored different frameworks of imitation learning. This includes new observational spaces such as those with history \cite{mandlekarWhatMattersLearning2022}, a large range of training procedures \cite{rossReductionImitationLearning2011, chiDiffusionPolicyVisuomotor2023a, florenceImplicitBehavioralCloning2022}, algorithms that target different operation assumptions \cite{rossReductionImitationLearning2011, liu2023modelbased}, and various policy architectures \cite{brohanRT1RoboticsTransformer2023, jangBCZZeroShotTask2022a, shafiullahBehaviorTransformersCloning2022a}. There have also been significant efforts to collect diverse, multimodal expert demonstrations \cite{brohanRT1RoboticsTransformer2023, ebertBridgeDataBoosting2022, walkeBridgeDataV2Dataset2024, collaborationOpenXEmbodimentRobotic2023}.
%Behavior-cloned policies have been extensively explored in the context of manipulation [CITE], locomotion [CITE, that quadruped paper], and others.
%%CF.1.31: BC is actually pretty rare in locomotion. You could mention driving though. There was a paper from NVIDIA around 2016 or so.
Recently, state-of-the-art performance has been achieved on models that predict long sequences of actions trained on diverse expert demonstrations \cite{zhaoAloha, chiDiffusionPolicyVisuomotor2023a}. We use one of these models, Diffusion Policy, in our experiments as the base policy that captures a strategy repertoire. However, an expressive policy is not sufficient to attempt a new scenario successfully. If the policy gets into an out-of-distribution error state, it can fail to recover \cite{rossReductionImitationLearning2011, tuSampleComplexityStability2022}. Our method works on top of the base policy to catch these error states and recover from them before they become irrecoverable. 
%%CF.1.31: this seems like an overstatement of our method -- we don't detection such error states that early. Our failure detection model can handle OOD states, but our policy can't. The reset ends up being important for putting the policy back in distribution. [resolved]
%% SK: 1.31: Tried to reword to address
% all without the need for explicit retrial demonstrations. 

\noindent \textbf{Adapting Rapidly:} Many real-world situations require adapting to distribution shifts under short timeframes. When we have access to an expert, we can use this expert to intervene on mistakes and show the robot how to recover from them \cite{mazoureAcceleratingExplorationRepresentation2023a, laskeyDARTNoiseInjection2017a, kellyHGDAggerInteractiveImitation2019a}. Expert corrections during test time can be effective, but constant expert supervision may limit the practicality of such approaches. Therefore, our method does not rely on expert corrections for adaptation. 
% [Learning robot anomaly recovery skills]. 
% However, it can be impractical to assume access to experts constantly, particularly in a real-world environment. 
% Our method does not use expert supervision, but our strategy evaluation can be interpreted as a simulated expert that detects and intervenes on suboptimal attempts. 

For fast adaptation, agents can also leverage prior experiences in the environment. It can be beneficial to collect experiences that cover many diverse behaviors, which helps aggregate a robust suite of approaches for the test-time scenario \cite{eysenbachDiversityAllYou2018a, kumarOneSolutionNot2020a, derekAdaptableAgentPopulations2021}. With sufficient diversity, adaptation becomes a matter of selecting a working strategy rather than making a new one. Our method provides a way of selecting strategies during adaptation. 
% We do not assume access to environment interactions outside of expert demonstrations, but we do assume that the demonstrations cover a broad set of strategies that can be applied to a test-time scenario. 

If adaptive behaviors are present in the training data, it is possible to distill adaptive behaviors naturally through offline meta-learning \cite{duanRLFastReinforcement2016a, mishraSimpleNeuralAttentive2018, pongOfflineMetaReinforcementLearning2022, rakellyEfficientOffPolicyMetaReinforcement2019a}. However, assuming such a well-crafted data distribution remains constrictive. It can be hard to show a diverse range of adaptive behaviors, especially for a scenario that is not seen during training. Our method has a non-parametric skewing component that can accomplish rapid behavior adaptation without requiring adaptive behavior in the training data. 
\noindent \textbf{Adapting Without Supervision:} As discussed above, most adaptation algorithms require some sort of supervision during adaptation. However, it can be impractical to assume access to experts constantly, particularly in a real-world environment. In our problem setup, we do not assume access to such supervision, including full environment resets. There have been works in the Single-Life RL domain with similar problem setups \cite{chenYouOnlyLive2022a}. These works have focused on self-supervised adaptation by learning an internal objective \cite{hansenSelfSupervisedPolicyAdaptation2021a} or searching for the best-pretrained policy \cite{chenAdaptOntheGoBehavior2023} for deployment. 
An important component of Single-Life RL approaches is the ability to detect and recover from failures. A common approach is to train a failure detection model that looks at the probability of success \cite{rodriguezAbortRetryGrasping}, other internal properties \cite{kuErrorDetectionSurprise2015, hejnaDistanceWeightedSupervised2023}, or an explicitly trained Q function for risky states \cite{thananjeyanRecoveryRLSafe2021a}.
Failure detection models can detect mistake states effectively, but the same out-of-distribution observations that degrade performance for the policy can also degrade performance for the failure detection models. We also create a failure detection model for our method, but critically, we rely on the model's self-consistency through a long history of past predictions. In the steps leading up to an out-of-distribution mistake, the predictions are still in-distribution. Therefore, the reliance on self-consistency is robust to distribution shifts in the robot's current behavior, allowing unexpected mistakes to be detected rapidly. 
% These past predictions are in-distribution, so the reliance on self-consistency is robust to distribution shifts in the robot's current behavior.

Single-Life RL approaches often use failure detection models in conjunction with a recovery policy that bring the robot back to an in-distribution state \cite{thananjeyanRecoveryRLSafe2021a, sharmaStateDistributionMatchingApproach2022a}. Our method also assumes access to a recovery policy and is compatible with such policies of any complexity. For our experiment setups, we find that a simple withdrawal of the robot arm is sufficient.

%%CF.1.31: "more likely to be in-distribution" is underselling I think.
%%CF.1.31: Annie's ROAM paper is also relevant.
%%CF.1.31: It could make sense to also have a paragraph on failure detection / anomaly detection in robotics.

\section{Background}

% Before presenting our approach, we describe two components that our method will build on: imitation learning with diffusion policy and Monte Carlo value estimation.

\subsection{Imitation Learning and Diffusion Policy}
\label{diffusion_policy}
We adopt an imitation learning framework, where we train a policy $\pi_\theta(a|s)$ to imitate expert behavior from a dataset. Specifically, we use a diffusion-based policy $\pi_\theta$ \cite{chiDiffusionPolicyVisuomotor2023a}. 
% In imitation learning, we train a policy $\pi_\theta(a|s)$ with parameters $\theta$ that maps from an observed state $s$ to an action $a$ using a dataset $\mathcal{D} :\{(s,a)\}$ of expert-level demonstrations. The goal is to fit $\pi_\theta(a | s)$ to match the expert distribution closely \cite{argallSurveyRobotLearning2009}. This paradigm can be used with many classes of $\pi_\theta$, but in our setup, we adopt the use of a diffusion-based policy \cite{chiDiffusionPolicyVisuomotor2023a}. 
%%CF.2.2: We predict the denoising steps, not the noising steps -- should make the language more precise. 
%%CF.2.2: Separately, it's a bit awkward to introduce notation x when we won't use it later at all. This is sort of a personal preference I guess, but could you introduce diffusion in the context of a conditional distribution a|s?
Diffusion is a paradigm for training generative models to predict a sequence of denoising steps $\epsilon_{1:n}$ that refines a noisy sample $a$ until it resembles a sample from the target distribution $p(a)$ \cite{hoDenoisingDiffusionProbabilistic2020a}. During test-time generation, we can sample $a \sim p(a)$ by starting from noise and predicting a sequence of denoising steps. 
To construct a diffusion \textit{policy}, the noise prediction model $\epsilon_\theta$ can be modified by conditioning it on the current state $s$. The final $\epsilon_\theta((a_1,..., a_k) | s)$ defines a policy and it allows us to sample a sequence of actions $\mathbf{a} = a_1, ..., a_k$ to execute open-loop in the environment. Predicting sequences of actions at one time allows for higher control frequencies and leads to higher performance \cite{chiDiffusionPolicyVisuomotor2023a, zhaoAloha}.

% During training, a diffusion model $\epsilon_\theta(x, k)$ takes a sample $a\sim D$ and minimizes a variant of the following objective:
% \begin{equation} 
% \label{diffusion}
% \mathcal{L}_{DF}(\theta) = E_{a \sim p(a), \epsilon_k \sim N(0, \sigma_k)}[\|\epsilon_\theta(a + \epsilon_k, k)- \epsilon_k\|^2]
% \end{equation}

% During test-time generation, we can sample a new $a$ by starting from noise and iteratively denoising using noise estimates from the trained $\epsilon_\theta$ and Langevin dynamics. There are various theoretical and empirical benefits to using a diffusion process for generating samples from $p(a)$ \cite{hoDenoisingDiffusionProbabilistic2020a}, and these benefits carry over to a generative model of form $\pi(a | s)$, which can be understood as a policy. 

% To construct a diffusion policy, the diffusion model can be modified by conditioning it on the current state $s$. The final $\epsilon_\theta((a_1,..., a_k) | s)$ allows us to sample a sequence of actions $\mathbf{a} = a_1, ..., a_k$ to execute open-loop in the environment. Predicting sequences of actions at one time allows for higher control frequencies and leads to higher performance \cite{chiDiffusionPolicyVisuomotor2023a, zhaoAloha}

\subsection{Value Functions From Expert Demonstrations}
% \textbf{TODO 
% - in the value function paragraph it would be good to cite eg Christian Hartikainen's work on learned distances and maybe also actionable models by Chebotar et al. for learning value distances; I believe Vichtyr Pong also had related work?}

\label{valuedemo}
% In many reinforcement learning paradigms, it is assumed that the environment provides a reward signal. 

Many expert-collected trajectories may not contain reward annotations, but it is still possible to learn a value function if we add a sparse reward label. 
% of progress \cite{hartikainenDynamicalDistanceLearning2020, mazoureAcceleratingExplorationRepresentation2023a} and 
% reward \cite{fuVariationalInverseControl2018a}.
%%CF.2.2: The progress part doesn't need to come in here, since that's more pertaining to our method. I'd just refer to it as value function learning. (Also separately, FWIW, I don't think the citations here are that important. Citations are most important when they are needed to support some claim that is otherwise unsupported.)
In our case, we make the common assumption that the last state $s_T$ in an expert demonstration is a success state \cite{chebotarActionableModelsUnsupervised2021, eysenbachContrastiveLearningGoalConditioned2023, kumarPreTrainingRobotsOffline2023a}.
%%CF.2.2: could also add PTR by Anikait to this citation block as another example.
We therefore assign reward $r_+$ to the success state at time $T$ and $r_-$ to all other states. With this, we can train a value function $V_\phi$ on the offline data by using Monte Carlo policy evaluation objective: 
% , specifically using Monte Carlo policy evaluation. We take a sampled state $s_t$, the timestep $t$, and the trajectory length $T$ from a trajectory in $\mathcal{D}$, allowing us to formulate this objective:
\vspace{-0.5em}
\begin{equation}
\label{valuefunction}
\mathcal{L}_\text{MC}(\phi) = E_{(t, T, s_t) \sim D}[(V_\phi(s_t)-( \gamma^{T - t}r_+ + \sum_{i = t}^T \gamma^{i - t}r_-))^2]
\end{equation}
When trained on demonstration data, the estimated value function should be a good approximation of both $V^{\pi^*}$ and $V^{\pi_\theta}$, where $\pi^*$ is the expert policy that collected the data and $\pi_\theta$ is a policy trained to imitate the demonstrations.

% When trained on the same data as an imitation learning policy $\pi_\theta$,
%%CF.2.2: Suggested rephrase that I think is clearer: When trained on demonstration data, the estimated value function should be a good approximation of both $V^{\pi^*}$ and $V^{\pi_\theta}$, where $\pi^*$$ is the expert policy that collected the data and $\pi_\theta$ is a policy trained to imitate the demonstrations.
% the value function $V_\phi$ should be a good approximation of the true $V^{\pi_\theta}$ for all $s \sim p_\mathcal{D}(s)$. 
%%CF.2.2: I would leave this next sentence to the methods section. Keep in mind that the reader might skip this section if they feel like they are already familiar with diffusion policy and MC value function estimation. The purpose of this section is generally just to get them up to speed with any background material they need to understand the methods section, and to cleanly separate out anything that is not our contribution.
% However, when $\pi_\theta$ is evaluated in a slightly different environment, the predicted $V_\phi$ may differ from the true performance of $\pi_\theta$. We can use this difference as a critical signal for monitoring the optimality of our strategy. 

\section{Rapid Trial and Error With Value Functions}
\subsection{Problem Setup}
In this work, we consider the problem setting of adapting to a novel scenario, where partial observability may necessitate interaction with the environment before it can be solved successfully. 
% For example, in a grasping environment, this information can include the valid affordances of an object that has never been seen before. A robot would need to discover these valid affordances by making mistakes on incorrect affordances, and then the robot can exploit its newfound knowledge to grab the object on a valid affordance.
Formally, we consider a Hidden-Parameter Markov Decision Process (HiP-MDP) \cite{killian2017robust}, defined by the tuple $(\mathcal{S}, \mathcal{A}, \mathcal{Z}, p_s(s_0), p_z(z), p(s'|s, a, z), r(s, a, z))$. The hidden variable $z$ influences the dynamics and the reward of the HiP-MDP, and the $z$ is sampled from $p(z)$ at the start of every episode. 
During training, we are given a dataset $\mathcal{D}$ that contains expert demonstrations in the HiP-MDPs on samples $z_1, ..., z_k \sim p_z(\cdot)$ of the hidden variables. These experts may already have privileged knowledge of the hidden variables, so they may not reliably show how to discover them. They do, however, reliably show how to solve the HiP-MDP once these parameters are known. During test-time, we sample $z' \sim p_z(\cdot)$ and we try to solve this HiP-MDP. To solve the $z'$ scenario, we need to uncover the hidden variable by interacting in the HiP-MDP. After we gain enough knowledge about the hidden parameter, we can then solve the $z'$ scenario by using the strategies learned from $\mathcal{D}$. 
% We are not concerned with extracting the exact $z$, but rather, systematically interact with the nesolve $\mathcal{T}'$ as quickly as possible. 

\subsection{Overview}
In \method, we propose an approach that allows robots to iterate quickly and systematically across different strategies to solve the HiP-MDP with a new $z' \sim p_z(\cdot)$. 
%%CF.2.2: similar to the above "solve a novel scenario" is going to sound like a huge overstatement to some people
%%CF.2.2: also be careful with using the word skill, since people can easily misinterpret that as programming skills into the model. I think "strategies" is a bit better, though perhaps not perfect
% Critically, we accomplish this goal without requiring expert interventions or additional training data. 
We first train a diverse base policy $\pi_\theta$ via imitation learning on an expert dataset (Section \ref{diffusion_policy}). During a novel scenario, we sample strategies from $\pi_\theta$ with special oversight. 
Our key insight is that we can estimate how efficiently our $\pi_\theta$ should solve this scenario, and therefore we can see if the policy's current strategy is suboptimal. 
We estimate this performance expectation using a value function $V_\phi$ trained on the same expert demonstrations used to make $\pi_\theta$ (Section \ref{faildetect}).
When we detect significant suboptimality with the current strategy, we recover and try a new strategy from $\pi_\theta$ while avoiding past mistakes through a skewed sampling approach (Section \ref{skew_method}). This trial-and-error process implicitly discovers parts of the hidden parameter $z'$, allowing us to solve the MDP quickly.
See Algorithm \ref{alg} for a summary of our method.

\subsection{Strategy Evaluation With Value Functions}
\label{faildetect}
In the first component of our method, we are interested in leveraging a trained value function $V_\phi$ to persistently evaluate the strategy that the base policy $\pi_\theta$ is currently executing. To evaluate progress, let us consider the Bellman target of $V_\phi(s_{t-1})$ computed with the trajectory of the current strategy:

% While applying $\pi_\theta$ to a new scenario with hidden variable $z'$, one might consider fine-tuning the policy and the value function through online interaction in $z'$. We want to minimize the number of interactions in the $z'$ scenario, which means that we want to avoid gradient-based tuning. Still, let us consider what happens when we compute the fine-tuning objective for $V_\phi$ with a transition $(s, a, r, s')$ from an interaction in the $z'$ scenario. The Bellman target is constructed as follows, using one sample as an approximation for the expectation
\vspace{-1em}
\begin{equation} 
y = r + \gamma E_{s_t}[V_\phi(s_t)] \approx r + \gamma V_\phi(s_t)
\end{equation}
If $V_\phi$ were perfectly accurate and $\pi_\theta$ perfectly following expert behavior, then $V_\phi(s_{t-1}) = E[r + V_\phi(s_t)]$. But novel $z'$ scenarios can create new and unexpected mistakes during a trajectory. 
% But because $\pi_\theta$ is a fitted function (Section \ref{diffusion_policy}) . 
Therefore, the Bellman error $V_\phi(s) - y$ may be non-zero for the current trajectory.

Specifically, let us consider the case where  $V_\phi(s_{t-1}) - y > 0$. Here, $V_\phi(s_{t-1})$ is overestimating the true value of $s_{t-1}$ in the novel $z'$ scenario. For example, suppose that the robot is attempting to grasp an object and sees a handle-like protrusion in $s_{t-1}$. We sample a strategy from $\pi_\theta$ that tries to pinch the handle. Unknown to $\pi_\theta$ at the time, this affordance in the novel $z'$ scenario is actually very slippery. After executing action $a \sim \pi_\theta$, the robot finds itself in a bad state $s_t$ without any grasp. Therefore, the value associated with $s_{t-1}$ should have been lower under this strategy, meaning the original handle grasping strategy was suboptimal.

% Because $V_\phi$ was trained on expert data that excludes the $z'$ scenario, the predicted $V_\phi(s)$ would have been an overestimate of the true value at that slippery handle state

From the analysis above, we argue that we can detect strategy suboptimality by comparing a value estimate $V_\phi(s_{t-1})$ with its bellman target $y = r + \gamma V_\phi(s_t)$. \textbf{If $V_\phi(s_{t-1})- y>0$, the executed strategy was suboptimal and the robot should recover and try a different strategy.}

% Let us consider the two cases of non-zero Bellman error. First, if $V_\phi(s) - y < 0$, it means that the true value of $s$ under $\pi_\theta$ in the $z'$ scenario is higher than predicted from expert data. Assuming fair generalization of $V_\phi$, this means that $\pi_\theta$ is performing comparably or better than an expert in this $z'$ scenario. 

% If the expert dataset is large enough, we can assume that this first case happens the majority of the time, as most skills can generalize in $\mathcal{T'}$. 

% From the analysis above, we present a straightforward way of evaluating the robot's current strategy at time $t$ using a pretrained value function. 

% \vspace{-0.5em}
% \begin{enumerate} 
% \item Extract the latest observed transition from the current interaction $(s_{t-1}, a_{t-1}, r_{t-1}, s_t)$ 
% %%CF.2.2: this sounds a little funny - convey that this the latest observed transition. It could be most helpful to refer to time t in the previous sentence and then denote this as (s_{t-1}, a_{t-1}, r_{t-1}, s_t). [and then propagate that change to the other steps] Or are you suggesting that you do this for multiple transitions and average the Bellman error?
% \item Compute the one-sample estimate of the Bellman target $y = r + \gamma V_\phi(s_t)$.
% \item Compare this target $y$ to $V_\phi(s_{t-1})$. If $V_\phi(s_{t-1}) > y$, then the current strategy is suboptimal. 
% \end{enumerate}

In practice, at timestep $t$, we have access to transitions from $0 \rightarrow t$. We can take advantage of this access by computing the lower-bias $k$-step returns for the Bellman target of $V_\phi(s_{t-k})$.
% \begin{equation}\label{bellman}y = \gamma^k V_\phi(s_{t}) + \sum_{i}^{k} \gamma^i r_{t -k + i}\end{equation}
Computing the multi-step returns also improves the robustness of our method against out-of-distribution mistake states. Even if $s_t$ is out of distribution, it is likely that $s_{t-k}$ was still in distribution for large enough $k$. Therefore, we are comparing the $V_\phi(s_t)$ to a reliable, in-distribution value estimation. 
% Unless the out-of-distribution state $V_\phi(s_t)$ behaves perfectly consistently with previous estimates, we will be able to detect an abnormality. 
Empirically, we observe that out-of-distribution mistakes cause drastic inconsistencies in $V_\phi(s_t)$ that are easily detectable through our method (Figure \ref{fig:exp1}). 

The above framework should be functional for any value function on any reward structure, including sparse rewards. 
% Our problem statement does not assume any external reward signal, but in Section \ref{valuedemo}, we described assigning a sparse, posthoc reward structure by labeling expert trajectories. While performing online interactions in the $z'$ scenario, we can assume that the current robot state is not a success state (otherwise we have solved the scenario and are done) and assign it a non-success reward.
The cleanest formulation happens when we use the reward structure described in Section \ref{valuedemo} and assign $r_- = -1$ to non-success states and $r_+ = 0$ to the final success state. If we also set $\gamma = 1$, our strategy evaluation framework with $k$-step returns becomes the following: 
\vspace{-0.5em}
\begin{equation} 
\label{progress}
V_\phi(s_{t - k}) \stackrel{?}{>} V_\phi(s_t) - k 
\end{equation}
The plot in Figure \ref{fig:exp1} shows the evolution $V_\phi(s_t) - k - V_\phi(s_{t-k})$ as the robot attempts a scenario multiple times. Each of the times that the expression crosses zero corresponds closely to a mistake in the environment. 

\subsection{Non-Parametric Behavior Skewing}
\label{skew_method}
After detecting a suboptimal strategy using our strategy evaluation framework (Section \ref{faildetect}), we can recover and try again. For manipulation tasks, recovery can be as simple as lifting the end-effector away from the object. Our proposed method does not set restrictions on recovery behavior, and the recovery can easily be a more complicated behavior. 

% With our strategy evaluation framework, we are able to detect whether the current strategy is suboptimal in the $z'$ scenario. When suboptimality happens, we need to recover to try another strategy. When we recover, we need to move the robot back into a state where $\pi_\theta(a| s)$ has sufficient diversity in potential strategies. For manipulation tasks, this often means lifting the robot end-effector sufficiently away from the object, allowing another affordance to be selected. Our proposed method does not set restrictions on recovery behavior, and the recovery can easily be a more complicated behavior. 

\begin{wrapfigure}{r}{0.45\textwidth}
\begin{minipage}{0.45\textwidth}
\vspace{-2.5em}
\begin{algorithm}[H]
\caption{\method (Novel Scenario Deployment)}
\begin{algorithmic}[1]
\label{alg}
% \footnotesize
\STATE $S^{avoid} \leftarrow \emptyset$
\WHILE{not successful}
        \STATE Sample $\mathbf{a}_1, ..., \mathbf{a}_k \sim \pi_\theta(\cdot | s_{t-1})$
        \STATE Select $\mathbf{a}^*$ according to Eq \ref{skew}
        \STATE Execute $\mathbf{a}^*$ in environment, get $s_t$
        \STATE Compute $k$-step Bellman target $y \leftarrow V_\phi(s_t) + \sum_{m=0}^{k-1} \gamma^m r_{t-k+m}$ 
        \IF{$V_\phi(s_{t-k}) > y$ }
            \STATE Recover robot to neutral state
            \STATE $S^{avoid} \leftarrow S^{avoid} \cup s_{t-1}$
        \ENDIF
\ENDWHILE
\end{algorithmic}
\end{algorithm}
\vspace{-2em}
\end{minipage}
\end{wrapfigure}

After recovery, we must replan using $\pi_\theta$. When we detect that a strategy is suboptimal, we also know the state $s^-$ that triggered this detection. Therefore, in future runs of $\pi_\theta$, we want to avoid $s^-$. To continue the previous example, if the $z'$ scenario was a grasping task with a somewhat slippery object, $s^-$ might correspond to being close to a slippery handle. Future samples from $\pi_\theta$ should avoid trying this slippery handle. 

One explicit way to avoid $s^-$ is adding a bias to the sampling from $\pi_\theta$. If we had access to a transition model $p(s' | s, a)$ and multiple samples $a_1, ..., a_k \sim \pi_\theta(\cdot | s)$, we could pick the sample $a^*$ such that $p(s^- | s, a^*)$ is lowest. This biased sampling method is non-parametric, meaning that we can modify $\pi_\theta$ with as few as a single $s^-$ avoidance point. Biased sampling is less expressive than a gradient-based update on $\pi_\theta(a | s)$, but as long as $\pi_\theta( a | s)$ has a sufficient variance, this biased sampling method can modify the action distribution significantly, achieving strategy adaptation. 

Since we do not have access to a transition model, we approximate the above approach. We use a diffusion policy as $\pi_\theta$, which outputs a sequence of actions $\mathbf{a}$ in the space of robot proprioception. We record $s^-$ in the space of proprioception ($s^-_{prop}$) and skew $\pi_\theta$ based on the distance of a sampled action from a set $S^{avoid}$ of avoidance points $\mathbf{s}^-_{prop}$. Formally, we sample $k$ action sequences $\mathbf{a}_1, ..., \mathbf{a}_k$ from $\pi_\theta$ and then select skewed action sequences according to the following:
% \begin{equation}
%     \label{skew}a^* = \min_{a_i \in \{1, ..., k\}} p(s^- | s, a_i) \approx\max_{a_i \in \{1, ..., k\}} ||s_{prop}^- - a_i||_2
% \end{equation}
\vspace{-0.4em}
\begin{equation}
    \label{skew}a^* = \arg \max_{\mathbf{a}_i \in \{1, ..., k\}} \min_{t, j}\|S^{avoid}_j - (\mathbf{a}_i)_t\|_2^2
\end{equation}
\vspace{-0.1em}
%%CF.2.2: The action sequence part doesn't come into play in this equation and it should. We should also introduce notation that differentiates an action from an action sequence (could potentially use \mathbf{a} for the sequence? I tentatively implemented that above, but need to implement the equation.
The expression of states as proprioception has some drawbacks, including reduced efficacy in dynamic scenes. However, it allows us to avoid training $p(s' | s, a)$, which is advantageous. 

In summary, our method takes a base robot policy and monitors its progress as it attempts a novel scenario (Section \ref{faildetect}). If we detect suboptimality, we recover and replan by skewing the base policy away from states that have caused suboptimality in the past. Algorithm \ref{alg} presents a summary of our novel scenario adaptation.

\section{Experiments}
We conduct a series of experiments designed to evaluate whether \method can boost the performance of a trained base policy. First, we qualitatively assess if our strategy evaluation method is making reasonable judgments about strategy optimality (Section \ref{suboptexp}). Next, we look at how \method boosts performance in simulated and real tasks (Section \ref{improvement}). Finally, we perform ablations of our method (Section \ref{improvement}).

\begin{wrapfigure}{r}{0.45\textwidth}
\vspace{-3.3em}
% \begin{figure}[t]
    \begin{center}
    \includegraphics[width = 0.45\textwidth]{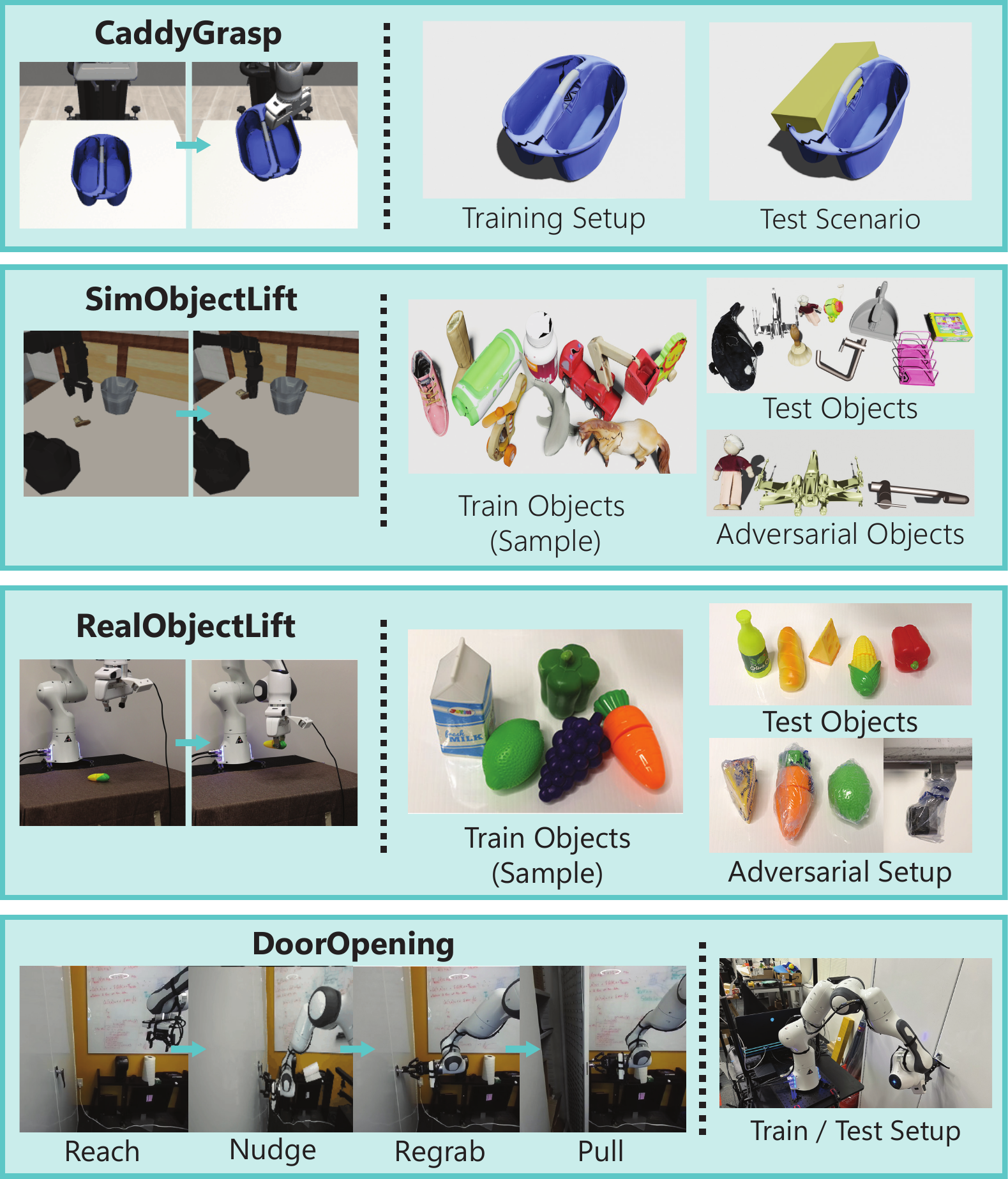}
    \end{center}
    \vspace{-0.8em}
    \caption{\small \textbf{Experiment Environments.} We consider four sets of experiment domains spanning simulated and real tasks of varying difficulty. } 
    % \vspace{-0.4cm}
    \label{fig:envs} 
% \end{figure}
\vspace{-4em}
\end{wrapfigure}

We train $\pi_\theta$ and $V_\phi$ on sets of human-teleoperated demonstrations. 
We make the design decision to represent $V_\phi$ as a categorical distribution, inspired by accuracy improvements seen in Distributional RL \cite{bellemareDistributionalPerspectiveReinforcement2017a, eysenbachSearchReplayBuffer2019}. We also adopt the reward structure of $r = -1$ for non-success states, which allows us to evaluate the current strategy by the simple formulation in Equation \ref{progress}. The distributional nature of $V_\phi$ allows us to turn the inequality of Equation \ref{progress} into a statistical comparison, which allows us to account for uncertainty in the estimated value. See the Appendix for more details.

\subsection{Experimental Domains}

We construct three main experimental domains that allow us to collect expert data on a set of training scenarios and test on a novel scenario. We also add a fourth domain to demonstrate that \method is robust to different types of tasks. 

In simulation, we introduce \textbf{CaddyGrasp} and \textbf{SimObjectLift}. The \textbf{CaddyGrasp} is a modified Robosuite \cite{robosuite2020} environment that requires the robot to lift a large shower caddy. The training set shows different grasps on all affordances of the caddy. During testing, an invisible obstacle (made visible in Figure \ref{fig:envs}) introduces a novel situation by blocking some of the affordances on the caddy, reducing the set of valid strategies. The \textbf{SimObjectLift} environment requires the robot to grab and lift a variety of objects from the Gazebo and ShapeNet datasets \cite{shapenet2015}. The training set shows grasps on a collection of objects. During testing, we provide novel situations by introducing held-out objects to the robot. We also test on an \textit{adversarial} set of difficult-to-grasp objects.  

On a real Franka Robot arm, we introduce \textbf{RealObjectLift} and \textbf{DoorOpening}. The expert data comes from human teleoperation with hand-held Oculus VR controllers. The \textbf{RealObjectLift} is the analogous task to \textbf{SimObjectLift} with real play kitchen objects that require a variety of strategies to lift. For \textbf{RealObjectLift}, we also test on an \textit{adversarial} setup with low-friction film taped to the objects and the robot grippers, which reduces the number of viable grasp strategies.  The additional long-horizon \textbf{DoorOpening} task requires the robot to grab, twist, and pull open a tool cabinet. This task shows the ability of \method to work beyond object-picking applications.

\subsection{When does \method Detect Suboptmality?}
\label{suboptexp}
\begin{wrapfigure}{r}{0.5\textwidth}
    \vspace{-2em}
    \centering %centers everything 
    \includegraphics[width = 0.49\textwidth]{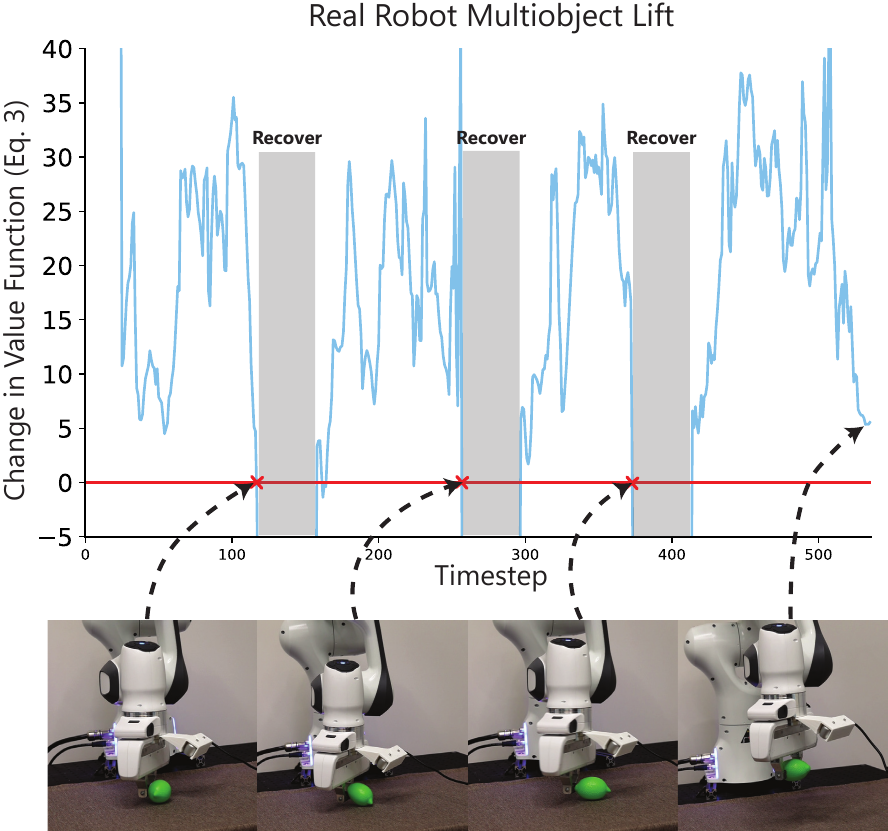}
    \caption{\small \textbf{Visualizing Suboptimality Detection. } The change in value function (Eq \ref{progress}) drops below zero immediately as the object slips out of the robot's grasp.} 
    \vspace{-3.5em}
    \label{fig:exp1} 
\end{wrapfigure}

For our first experiment, we qualitatively analyze how \method detects suboptimality by querying the value function. In Figure \ref{fig:exp1}, we plot $V_\phi(s_t)  - V_\phi(s_{t-k})- k$ over a trajectory. As detailed in Section \ref{faildetect}, if this expression drops below zero, it indicates a suboptimality. In the figure, we see that the expression crosses the threshold three times, each happening a few frames after the plastic lime slips out of the robot's grasp. These results are representative of a general property of our method: we can detect mistakes quickly, allowing the robot to save time during adaptation and also prevent it from reaching unrecoverable error states.

\subsection{Does \method Improve Performance Of Expert-Trained Policies?}

\begin{figure*}[tb]
    \centering %centers everything 
    \includegraphics[width = 0.99\textwidth]{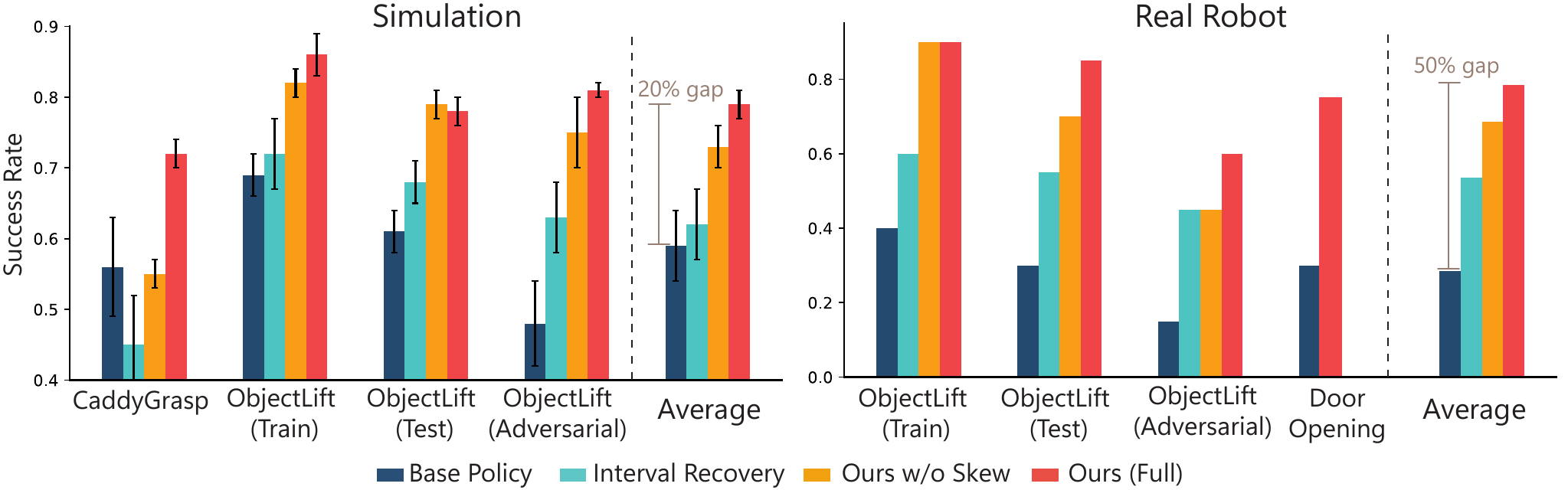}
    \vspace{-0.5em}
    \caption{\small \textbf{Comparing \method To Relevant Baselines.} In both simulation and real robots, our method boosts the performance of the base policy and outperforms an interval recovery baseline. Note that the DoorOpening results are limited because we lost the environment prematurely (Appendix \ref{addl_exp_robot}).} 
    \label{fig:exp3} 
     \vspace{-2em}
\end{figure*}

For our main set of quantitative experiments, we want to see how our method performs against relevant baselines. We test the expert-trained $\pi_\theta$ without our method, and then we use the same $\pi_\theta$ with our method by adding the suboptimality detection and skewed sampling. We test our method without the skewed sampling as an ablation, and we also test our method against an \textit{interval recovery} baseline, where we trigger recoveries regularly based on the average number of steps it takes an expert to finish the task. For simulation, we compute success rates and standard deviations over 100 trials across three versions of $\pi_\theta, V_\phi$ trained with different random seeds. For the real robot experiment, we conduct 20 trials for each result. In all setups, we use a matched-pair design that ensures all algorithms get the same set of objects and starting orientations. 

\label{improvement}
As seen in Figure \ref{fig:exp3}, \method improves performance significantly over baselines in the three main real and simulated domains. The \textbf{CaddyGrasp} environment shows an especially large jump between the base policy and our method, as the base policy often gets stuck trying to grab a part of the caddy that has been blocked. \textbf{CaddyGrasp} also shows a large jump in performance when we add skewed sampling. Without skewed sampling, the robot often tries the blocked regions over and over. The interval reset baseline also exhibits this failure mode. This \textbf{CaddyGrasp} environment is somewhat contrived, but it demonstrates clearly and intuitively the benefits of both components of our method. Appendix \ref{skewexp} shows an additional \textbf{CaddyGrasp} visualization of skewed sampling.

The \textbf{SimObjectLift} environment also shows a significant difference between our method and baselines. We test performance on the train objects, held-out test objects, and adversarially-chosen objects. The more difficult situations (adversarial) created more opportunities for mistakes and recoveries, which increased the performance gap between the bare base policy and our method. Qualitatively, the main failure mode of the base policy is a failed grasp. Without our method, the base policy will hover over the dropped object with its grippers closed. In contrast, our method can quickly detect this mistake, recover, and retry. We also observe instances of early recovery, where our method will trigger a recovery before the robot attempts to lift on a bad grasp, which reduces the time per trial and the risk of irrecoverable changing the environment (e.g. knocking the object off the table). Our method still improved performance on the non-novel train set objects, which demonstrates that mistake detection is broadly useful. On average, our full method improves the base policy performance by $20\%$ (Figure \ref{fig:exp3}). See Appendix \ref{simobjectlift_bonus} for more details.

The \textbf{RealObjectLift} environment shows similar trends as its analogous simulated task. The common failure mode of the base policy is a misgrasp, where the object shifts out of the grippers as they close. The base policy will attempt to lift and end up in an error state. We exacerbate this failure mode in the \textit{adversarial} setup with low-friction film. This failure mode is readily detected by our method, and the skewed sampling encourages the robot to try different grasp point after recovery. On average, our full method improves base policy performance by $50\%$ (Figure \ref{fig:exp3}).  We introduce an additional, difficult \textbf{DoorOpening} task and evaluate our full method against the bare base policy. Figure \ref{fig:exp3} shows that our full method boosts base policy performance by more than two times. Commonly, the base policy pushes the handle of the door too close to the pivot, which gets the robot stuck. Our method will detect this mistake, withdraw the arm and try again. In theory, the \method framework will work for any type of task, and \textbf{DoorOpening} supports this notion. See the Appendix \ref{realobjectlift_bonus}, \ref{addl_exp_robot} for more details about \textbf{RealObjectLift} and \textbf{DoorOpening} results, respectively.

The main experiment results also include two critical ablations. First, the interval recovery baseline ablates the  strategy evaluation afforded by our method. The reduction in performance shows that the statistics of an expert (average completion time) is not sufficient to determine when to recover from a mistake. Second, \textbf{Ours w/o Skew} ablation shows that the skewed resampling can boost performance by avoiding bad grasps, although it contributes less to overall performance than strategy evaluation. 
See the Appendix \ref{ablation} for more details.

\section{Conclusion}
 We have presented \method, a method for adapting quickly to a novel scenario by enhancing the performance of a base policy. We monitor the viability of a strategy by using the self-consistency of a trained value function.  We trigger a robot recovery when we detect suboptimal progress, and we also skew the robot policy away from past suboptimal states. Our experiments show that \method can boost success rates by more than 20\% in simulation and 50\% on a real robot. 
 % Our method can also potentially work on top of many pre-trained policies, and it requires no additional data, nor does it require expert supervision or fine-tuning to adapt quickly.

\noindent \textbf{Limitations and Future Work.}
We are excited by the possibilities of \method, but some limitations remain. Because we do not modify the expert-trained policy, our method will not imbue the robot with any strategies that it does not already have. Also, the framework of \method assumes access to a robust recovery policy, which may not be trivial for more involved tasks. Finally, we used distance in proprioceptive space for skewing, but such a metric would not work well if the environment were very dynamic. In future work, it is worth exploring a more expressive skewing method that operates in the robot's \textit{strategy space}, much like how a human may explicitly try conceptually different strategies. 

\bibliography{references}

\begin{thebibliography}{49}
\providecommand{\natexlab}[1]{#1}
\providecommand{\url}[1]{\texttt{#1}}
\expandafter\ifx\csname urlstyle\endcsname\relax
  \providecommand{\doi}[1]{doi: #1}\else
  \providecommand{\doi}{doi: \begingroup \urlstyle{rm}\Url}\fi

\bibitem[Zhao et~al.(2023)Zhao, Kumar, Levine, and Finn]{zhaoAloha}
T.~Zhao, V.~Kumar, S.~Levine, and C.~Finn.
\newblock Learning fine-grained bimanual manipulation with low-cost hardware.
\newblock \emph{Robotics: Science and Systems}, 2023.
\newblock URL \url{https://www.roboticsproceedings.org/rss19/p016.pdf}.

\bibitem[Grannen et~al.()Grannen, Wu, Vu, and Sadigh]{grannenStabilizeActLearning2023}
J.~Grannen, Y.~Wu, B.~Vu, and D.~Sadigh.
\newblock Stabilize to {{Act}}: {{Learning}} to {{Coordinate}} for {{Bimanual Manipulation}}.
\newblock URL \url{http://arxiv.org/abs/2309.01087}.

\bibitem[Xie et~al.()Xie, Chowdhury, De~Paolis~Kaluza, Zhao, Wong, and Yu]{xieDeepImitationLearning2020}
F.~Xie, A.~Chowdhury, M.~C. De~Paolis~Kaluza, L.~Zhao, L.~Wong, and R.~Yu.
\newblock Deep {{Imitation Learning}} for {{Bimanual Robotic Manipulation}}.
\newblock In \emph{Advances in {{Neural Information Processing Systems}}}, volume~33, pages 2327--2337. {Curran Associates, Inc.}
\newblock URL \url{https://proceedings.neurips.cc/paper/2020/hash/18a010d2a9813e91907ce88cd9143fdf-Abstract.html}.

\bibitem[Smith et~al.()Smith, Karayiannidis, Nalpantidis, Gratal, Qi, Dimarogonas, and Kragic]{smithDualArmManipulation2012}
C.~Smith, Y.~Karayiannidis, L.~Nalpantidis, X.~Gratal, P.~Qi, D.~V. Dimarogonas, and D.~Kragic.
\newblock Dual arm manipulation—{{A}} survey.
\newblock 60\penalty0 (10):\penalty0 1340--1353.
\newblock ISSN 0921-8890.
\newblock \doi{10.1016/j.robot.2012.07.005}.
\newblock URL \url{https://www.sciencedirect.com/science/article/pii/S092188901200108X}.

\bibitem[Schaal()]{schaalImitationLearningRoute1999}
S.~Schaal.
\newblock Is imitation learning the route to humanoid robots?
\newblock 3\penalty0 (6):\penalty0 233--242.
\newblock ISSN 1364-6613.
\newblock \doi{10.1016/S1364-6613(99)01327-3}.
\newblock URL \url{https://www.sciencedirect.com/science/article/pii/S1364661399013273}.

\bibitem[Ross et~al.()Ross, Gordon, and Bagnell]{rossReductionImitationLearning2011}
S.~Ross, G.~Gordon, and D.~Bagnell.
\newblock A {{Reduction}} of {{Imitation Learning}} and {{Structured Prediction}} to {{No-Regret Online Learning}}.
\newblock In \emph{Proceedings of the {{Fourteenth International Conference}} on {{Artificial Intelligence}} and {{Statistics}}}, pages 627--635. {JMLR Workshop and Conference Proceedings}.
\newblock URL \url{https://proceedings.mlr.press/v15/ross11a.html}.

\bibitem[Tu et~al.()Tu, Robey, Zhang, and Matni]{tuSampleComplexityStability2022}
S.~Tu, A.~Robey, T.~Zhang, and N.~Matni.
\newblock On the {{Sample Complexity}} of {{Stability Constrained Imitation Learning}}.
\newblock In \emph{Proceedings of {{The}} 4th {{Annual Learning}} for {{Dynamics}} and {{Control Conference}}}, pages 180--191. {PMLR}.
\newblock URL \url{https://proceedings.mlr.press/v168/tu22a.html}.

\bibitem[Mazoure et~al.()Mazoure, Bruce, Precup, Fergus, and Anand]{mazoureAcceleratingExplorationRepresentation2023a}
B.~Mazoure, J.~Bruce, D.~Precup, R.~Fergus, and A.~Anand.
\newblock Accelerating exploration and representation learning with offline pre-training.
\newblock URL \url{http://arxiv.org/abs/2304.00046}.

\bibitem[Pomerleau(1989)]{alvinn}
D.~Pomerleau.
\newblock Alvinn: An autonomous land vehicle in a neural network.
\newblock In D.~Touretzky, editor, \emph{Proceedings of (NeurIPS) Neural Information Processing Systems}, pages 305 -- 313. Morgan Kaufmann, December 1989.

\bibitem[Argall et~al.()Argall, Chernova, Veloso, and Browning]{argallSurveyRobotLearning2009}
B.~D. Argall, S.~Chernova, M.~Veloso, and B.~Browning.
\newblock A survey of robot learning from demonstration.
\newblock 57\penalty0 (5):\penalty0 469--483.
\newblock ISSN 09218890.
\newblock \doi{10.1016/j.robot.2008.10.024}.
\newblock URL \url{https://linkinghub.elsevier.com/retrieve/pii/S0921889008001772}.

\bibitem[Billard et~al.()Billard, Calinon, Dillmann, and Schaal]{billardRobotProgrammingDemonstration2008}
A.~Billard, S.~Calinon, R.~Dillmann, and S.~Schaal.
\newblock Robot {{Programming}} by {{Demonstration}}.
\newblock In B.~Siciliano and O.~Khatib, editors, \emph{Springer {{Handbook}} of {{Robotics}}}, pages 1371--1394. {Springer}.
\newblock ISBN 978-3-540-30301-5.
\newblock \doi{10.1007/978-3-540-30301-5_60}.
\newblock URL \url{https://doi.org/10.1007/978-3-540-30301-5_60}.

\bibitem[Schaal et~al.()Schaal, Ijspeert, and Billard]{schaalComputationalApproachesMotor2003}
S.~Schaal, A.~Ijspeert, and A.~Billard.
\newblock Computational approaches to motor learning by imitation.
\newblock 358\penalty0 (1431):\penalty0 537--547.
\newblock ISSN 0962-8436.
\newblock \doi{10.1098/rstb.2002.1258}.
\newblock URL \url{https://www.ncbi.nlm.nih.gov/pmc/articles/PMC1693137/}.

\bibitem[Mandlekar et~al.()Mandlekar, Xu, Wong, Nasiriany, Wang, Kulkarni, Fei-Fei, Savarese, Zhu, and Martín-Martín]{mandlekarWhatMattersLearning2022}
A.~Mandlekar, D.~Xu, J.~Wong, S.~Nasiriany, C.~Wang, R.~Kulkarni, L.~Fei-Fei, S.~Savarese, Y.~Zhu, and R.~Martín-Martín.
\newblock What {{Matters}} in {{Learning}} from {{Offline Human Demonstrations}} for {{Robot Manipulation}}.
\newblock In \emph{Proceedings of the 5th {{Conference}} on {{Robot Learning}}}, pages 1678--1690. {PMLR}.
\newblock URL \url{https://proceedings.mlr.press/v164/mandlekar22a.html}.

\bibitem[Chi et~al.()Chi, Feng, Du, Xu, Cousineau, Burchfiel, and Song]{chiDiffusionPolicyVisuomotor2023a}
C.~Chi, S.~Feng, Y.~Du, Z.~Xu, E.~Cousineau, B.~Burchfiel, and S.~Song.
\newblock Diffusion {{Policy}}: {{Visuomotor Policy Learning}} via {{Action Diffusion}}.
\newblock In \emph{Robotics: {{Science}} and {{Systems XIX}}}. {Robotics: Science and Systems Foundation}.
\newblock ISBN 978-0-9923747-9-2.
\newblock \doi{10.15607/RSS.2023.XIX.026}.
\newblock URL \url{http://www.roboticsproceedings.org/rss19/p026.pdf}.

\bibitem[Florence et~al.()Florence, Lynch, Zeng, Ramirez, Wahid, Downs, Wong, Lee, Mordatch, and Tompson]{florenceImplicitBehavioralCloning2022}
P.~Florence, C.~Lynch, A.~Zeng, O.~A. Ramirez, A.~Wahid, L.~Downs, A.~Wong, J.~Lee, I.~Mordatch, and J.~Tompson.
\newblock Implicit {{Behavioral Cloning}}.
\newblock In \emph{Proceedings of the 5th {{Conference}} on {{Robot Learning}}}, pages 158--168. {PMLR}.
\newblock URL \url{https://proceedings.mlr.press/v164/florence22a.html}.

\bibitem[Liu et~al.(2023)Liu, Dass, Martín-Martín, and Zhu]{liu2023modelbased}
H.~Liu, S.~Dass, R.~Martín-Martín, and Y.~Zhu.
\newblock Model-based runtime monitoring with interactive imitation learning, 2023.

\bibitem[Brohan et~al.()Brohan, Brown, Carbajal, Chebotar, Dabis, Finn, Gopalakrishnan, Hausman, Herzog, Hsu, Ibarz, Ichter, Irpan, Jackson, Jesmonth, Joshi, Julian, Kalashnikov, Kuang, Leal, Lee, Levine, Lu, Malla, Manjunath, Mordatch, Nachum, Parada, Peralta, Perez, Pertsch, Quiambao, Rao, Ryoo, Salazar, Sanketi, Sayed, Singh, Sontakke, Stone, Tan, Tran, Vanhoucke, Vega, Vuong, Xia, Xiao, Xu, Xu, Yu, and Zitkovich]{brohanRT1RoboticsTransformer2023}
A.~Brohan, N.~Brown, J.~Carbajal, Y.~Chebotar, J.~Dabis, C.~Finn, K.~Gopalakrishnan, K.~Hausman, A.~Herzog, J.~Hsu, J.~Ibarz, B.~Ichter, A.~Irpan, T.~Jackson, S.~Jesmonth, N.~Joshi, R.~Julian, D.~Kalashnikov, Y.~Kuang, I.~Leal, K.-H. Lee, S.~Levine, Y.~Lu, U.~Malla, D.~Manjunath, I.~Mordatch, O.~Nachum, C.~Parada, J.~Peralta, E.~Perez, K.~Pertsch, J.~Quiambao, K.~Rao, M.~Ryoo, G.~Salazar, P.~Sanketi, K.~Sayed, J.~Singh, S.~Sontakke, A.~Stone, C.~Tan, H.~Tran, V.~Vanhoucke, S.~Vega, Q.~Vuong, F.~Xia, T.~Xiao, P.~Xu, S.~Xu, T.~Yu, and B.~Zitkovich.
\newblock {{RT-1}}: {{Robotics Transformer}} for {{Real-World Control}} at {{Scale}}.
\newblock In \emph{Robotics: {{Science}} and {{Systems XIX}}}. {Robotics: Science and Systems Foundation}.
\newblock ISBN 978-0-9923747-9-2.
\newblock \doi{10.15607/RSS.2023.XIX.025}.
\newblock URL \url{http://www.roboticsproceedings.org/rss19/p025.pdf}.

\bibitem[Jang et~al.()Jang, Irpan, Khansari, Kappler, Ebert, Lynch, Levine, and Finn]{jangBCZZeroShotTask2022a}
E.~Jang, A.~Irpan, M.~Khansari, D.~Kappler, F.~Ebert, C.~Lynch, S.~Levine, and C.~Finn.
\newblock {{BC-Z}}: {{Zero-Shot Task Generalization}} with {{Robotic Imitation Learning}}.
\newblock In \emph{Proceedings of the 5th {{Conference}} on {{Robot Learning}}}, pages 991--1002. {PMLR}.
\newblock URL \url{https://proceedings.mlr.press/v164/jang22a.html}.

\bibitem[Shafiullah et~al.()Shafiullah, Cui, Altanzaya, and Pinto]{shafiullahBehaviorTransformersCloning2022a}
N.~M. Shafiullah, Z.~Cui, A.~A. Altanzaya, and L.~Pinto.
\newblock Behavior {{Transformers}}: {{Cloning}} \$k\$ modes with one stone.
\newblock 35:\penalty0 22955--22968.
\newblock URL \url{https://proceedings.neurips.cc/paper_files/paper/2022/hash/90d17e882adbdda42349db6f50123817-Abstract-Conference.html}.

\bibitem[Ebert et~al.()Ebert, Yang, Schmeckpeper, Bucher, Georgakis, Daniilidis, Finn, and Levine]{ebertBridgeDataBoosting2022}
F.~Ebert, Y.~Yang, K.~Schmeckpeper, B.~Bucher, G.~Georgakis, K.~Daniilidis, C.~Finn, and S.~Levine.
\newblock Bridge {{Data}}: {{Boosting Generalization}} of {{Robotic Skills}} with {{Cross-Domain Datasets}}.
\newblock In \emph{Robotics: {{Science}} and {{Systems XVIII}}}. {Robotics: Science and Systems Foundation}.
\newblock ISBN 978-0-9923747-8-5.
\newblock \doi{10.15607/RSS.2022.XVIII.063}.
\newblock URL \url{http://www.roboticsproceedings.org/rss18/p063.pdf}.

\bibitem[Walke et~al.()Walke, Black, Lee, Kim, Du, Zheng, Zhao, Hansen-Estruch, Vuong, He, Myers, Fang, Finn, and Levine]{walkeBridgeDataV2Dataset2024}
H.~Walke, K.~Black, A.~Lee, M.~J. Kim, M.~Du, C.~Zheng, T.~Zhao, P.~Hansen-Estruch, Q.~Vuong, A.~He, V.~Myers, K.~Fang, C.~Finn, and S.~Levine.
\newblock {{BridgeData V2}}: {{A Dataset}} for {{Robot Learning}} at {{Scale}}.
\newblock URL \url{http://arxiv.org/abs/2308.12952}.

\bibitem[Collaboration()]{collaborationOpenXEmbodimentRobotic2023}
O.~X.-E. Collaboration.
\newblock Open {{X-Embodiment}}: {{Robotic Learning Datasets}} and {{RT-X Models}}.
\newblock URL \url{http://arxiv.org/abs/2310.08864}.

\bibitem[Laskey et~al.()Laskey, Lee, Fox, Dragan, and Goldberg]{laskeyDARTNoiseInjection2017a}
M.~Laskey, J.~Lee, R.~Fox, A.~Dragan, and K.~Goldberg.
\newblock {{DART}}: {{Noise Injection}} for {{Robust Imitation Learning}}.
\newblock URL \url{http://arxiv.org/abs/1703.09327}.

\bibitem[Kelly et~al.()Kelly, Sidrane, Driggs-Campbell, and Kochenderfer]{kellyHGDAggerInteractiveImitation2019a}
M.~Kelly, C.~Sidrane, K.~Driggs-Campbell, and M.~J. Kochenderfer.
\newblock {{HG-DAgger}}: {{Interactive Imitation Learning}} with {{Human Experts}}.
\newblock URL \url{http://arxiv.org/abs/1810.02890}.

\bibitem[Eysenbach et~al.()Eysenbach, Gupta, Ibarz, and Levine]{eysenbachDiversityAllYou2018a}
B.~Eysenbach, A.~Gupta, J.~Ibarz, and S.~Levine.
\newblock Diversity is {{All You Need}}: {{Learning Skills}} without a {{Reward Function}}.
\newblock URL \url{http://arxiv.org/abs/1802.06070}.

\bibitem[Kumar et~al.()Kumar, Kumar, Levine, and Finn]{kumarOneSolutionNot2020a}
S.~Kumar, A.~Kumar, S.~Levine, and C.~Finn.
\newblock One {{Solution}} is {{Not All You Need}}: {{Few-Shot Extrapolation}} via {{Structured MaxEnt RL}}.
\newblock URL \url{http://arxiv.org/abs/2010.14484}.

\bibitem[Derek and Isola()]{derekAdaptableAgentPopulations2021}
K.~Derek and P.~Isola.
\newblock Adaptable {{Agent Populations}} via a {{Generative Model}} of {{Policies}}.
\newblock In \emph{Advances in {{Neural Information Processing Systems}}}, volume~34, pages 3902--3913. {Curran Associates, Inc.}
\newblock URL \url{https://proceedings.neurips.cc/paper/2021/hash/1fc8c3d03b0021478a8c9ebdcd457c67-Abstract.html}.

\bibitem[Duan et~al.()Duan, Schulman, Chen, Bartlett, Sutskever, and Abbeel]{duanRLFastReinforcement2016a}
Y.~Duan, J.~Schulman, X.~Chen, P.~L. Bartlett, I.~Sutskever, and P.~Abbeel.
\newblock {{RL}}\$\^2\$: {{Fast Reinforcement Learning}} via {{Slow Reinforcement Learning}}.
\newblock URL \url{http://arxiv.org/abs/1611.02779}.

\bibitem[Mishra et~al.()Mishra, Rohaninejad, Chen, and Abbeel]{mishraSimpleNeuralAttentive2018}
N.~Mishra, M.~Rohaninejad, X.~Chen, and P.~Abbeel.
\newblock A {{Simple Neural Attentive Meta-Learner}}.
\newblock URL \url{http://arxiv.org/abs/1707.03141}.

\bibitem[Pong et~al.()Pong, Nair, Smith, Huang, and Levine]{pongOfflineMetaReinforcementLearning2022}
V.~H. Pong, A.~Nair, L.~Smith, C.~Huang, and S.~Levine.
\newblock Offline {{Meta-Reinforcement Learning}} with {{Online Self-Supervision}}.
\newblock URL \url{http://arxiv.org/abs/2107.03974}.

\bibitem[Rakelly et~al.()Rakelly, Zhou, Quillen, Finn, and Levine]{rakellyEfficientOffPolicyMetaReinforcement2019a}
K.~Rakelly, A.~Zhou, D.~Quillen, C.~Finn, and S.~Levine.
\newblock Efficient {{Off-Policy Meta-Reinforcement Learning}} via {{Probabilistic Context Variables}}.
\newblock URL \url{http://arxiv.org/abs/1903.08254}.

\bibitem[Chen et~al.()Chen, Sharma, Levine, and Finn]{chenYouOnlyLive2022a}
A.~S. Chen, A.~Sharma, S.~Levine, and C.~Finn.
\newblock You {{Only Live Once}}: {{Single-Life Reinforcement Learning}}.
\newblock URL \url{http://arxiv.org/abs/2210.08863}.

\bibitem[Hansen et~al.()Hansen, Jangir, Sun, Alenyà, Abbeel, Efros, Pinto, and Wang]{hansenSelfSupervisedPolicyAdaptation2021a}
N.~Hansen, R.~Jangir, Y.~Sun, G.~Alenyà, P.~Abbeel, A.~A. Efros, L.~Pinto, and X.~Wang.
\newblock Self-{{Supervised Policy Adaptation}} during {{Deployment}}.
\newblock URL \url{http://arxiv.org/abs/2007.04309}.

\bibitem[Chen et~al.()Chen, Chada, Smith, Sharma, Fu, Levine, and Finn]{chenAdaptOntheGoBehavior2023}
A.~S. Chen, G.~Chada, L.~Smith, A.~Sharma, Z.~Fu, S.~Levine, and C.~Finn.
\newblock Adapt {{On-the-Go}}: {{Behavior Modulation}} for {{Single-Life Robot Deployment}}.
\newblock URL \url{http://arxiv.org/abs/2311.01059}.

\bibitem[Rodriguez et~al.()Rodriguez, Mason, Srinivasa, Bernstein, and Zirbel]{rodriguezAbortRetryGrasping}
A.~Rodriguez, M.~T. Mason, S.~Srinivasa, M.~Bernstein, and A.~Zirbel.
\newblock Abort and {{Retry}} in {{Grasping}}.

\bibitem[Ku et~al.()Ku, Ruiken, Learned-Miller, and Grupen]{kuErrorDetectionSurprise2015}
L.~Y. Ku, D.~Ruiken, E.~Learned-Miller, and R.~Grupen.
\newblock Error detection and surprise in stochastic robot actions.
\newblock In \emph{2015 {{IEEE-RAS}} 15th {{International Conference}} on {{Humanoid Robots}} ({{Humanoids}})}, pages 1096--1101.
\newblock \doi{10.1109/HUMANOIDS.2015.7363505}.
\newblock URL \url{https://ieeexplore.ieee.org/document/7363505}.

\bibitem[Hejna et~al.()Hejna, Gao, and Sadigh]{hejnaDistanceWeightedSupervised2023}
J.~Hejna, J.~Gao, and D.~Sadigh.
\newblock Distance {{Weighted Supervised Learning}} for {{Offline Interaction Data}}.
\newblock In \emph{Proceedings of the 40th {{International Conference}} on {{Machine Learning}}}, pages 12882--12906. PMLR.
\newblock URL \url{https://proceedings.mlr.press/v202/hejna23a.html}.

\bibitem[Thananjeyan et~al.()Thananjeyan, Balakrishna, Nair, Luo, Srinivasan, Hwang, Gonzalez, Ibarz, Finn, and Goldberg]{thananjeyanRecoveryRLSafe2021a}
B.~Thananjeyan, A.~Balakrishna, S.~Nair, M.~Luo, K.~Srinivasan, M.~Hwang, J.~E. Gonzalez, J.~Ibarz, C.~Finn, and K.~Goldberg.
\newblock Recovery {{RL}}: {{Safe Reinforcement Learning}} with {{Learned Recovery Zones}}.
\newblock URL \url{http://arxiv.org/abs/2010.15920}.

\bibitem[Sharma et~al.()Sharma, Ahmad, and Finn]{sharmaStateDistributionMatchingApproach2022a}
A.~Sharma, R.~Ahmad, and C.~Finn.
\newblock A {{State-Distribution Matching Approach}} to {{Non-Episodic Reinforcement Learning}}.
\newblock URL \url{http://arxiv.org/abs/2205.05212}.

\bibitem[Ho et~al.()Ho, Jain, and Abbeel]{hoDenoisingDiffusionProbabilistic2020a}
J.~Ho, A.~Jain, and P.~Abbeel.
\newblock Denoising {{Diffusion Probabilistic Models}}.
\newblock In \emph{Advances in {{Neural Information Processing Systems}}}, volume~33, pages 6840--6851. {Curran Associates, Inc.}
\newblock URL \url{https://proceedings.neurips.cc/paper/2020/hash/4c5bcfec8584af0d967f1ab10179ca4b-Abstract.html}.

\bibitem[Chebotar et~al.()Chebotar, Hausman, Lu, Xiao, Kalashnikov, Varley, Irpan, Eysenbach, Julian, Finn, and Levine]{chebotarActionableModelsUnsupervised2021}
Y.~Chebotar, K.~Hausman, Y.~Lu, T.~Xiao, D.~Kalashnikov, J.~Varley, A.~Irpan, B.~Eysenbach, R.~Julian, C.~Finn, and S.~Levine.
\newblock Actionable {{Models}}: {{Unsupervised Offline Reinforcement Learning}} of {{Robotic Skills}}.
\newblock URL \url{http://arxiv.org/abs/2104.07749}.

\bibitem[Eysenbach et~al.()Eysenbach, Zhang, Salakhutdinov, and Levine]{eysenbachContrastiveLearningGoalConditioned2023}
B.~Eysenbach, T.~Zhang, R.~Salakhutdinov, and S.~Levine.
\newblock Contrastive {{Learning}} as {{Goal-Conditioned Reinforcement Learning}}.
\newblock URL \url{http://arxiv.org/abs/2206.07568}.

\bibitem[Kumar et~al.()Kumar, Singh, Ebert, Nakamoto, Yang, Finn, and Levine]{kumarPreTrainingRobotsOffline2023a}
A.~Kumar, A.~Singh, F.~Ebert, M.~Nakamoto, Y.~Yang, C.~Finn, and S.~Levine.
\newblock Pre-{{Training}} for {{Robots}}: {{Offline RL Enables Learning New Tasks}} from a {{Handful}} of {{Trials}}.
\newblock URL \url{http://arxiv.org/abs/2210.05178}.

\bibitem[Killian et~al.(2017)Killian, Daulton, Konidaris, and Doshi-Velez]{killian2017robust}
T.~W. Killian, S.~Daulton, G.~Konidaris, and F.~Doshi-Velez.
\newblock Robust and efficient transfer learning with hidden parameter markov decision processes.
\newblock \emph{Advances in neural information processing systems}, 30, 2017.

\bibitem[Bellemare et~al.()Bellemare, Dabney, and Munos]{bellemareDistributionalPerspectiveReinforcement2017a}
M.~G. Bellemare, W.~Dabney, and R.~Munos.
\newblock A {{Distributional Perspective}} on {{Reinforcement Learning}}.
\newblock URL \url{http://arxiv.org/abs/1707.06887}.

\bibitem[Eysenbach et~al.()Eysenbach, Salakhutdinov, and Levine]{eysenbachSearchReplayBuffer2019}
B.~Eysenbach, R.~Salakhutdinov, and S.~Levine.
\newblock Search on the {{Replay Buffer}}: {{Bridging Planning}} and {{Reinforcement Learning}}.
\newblock URL \url{http://arxiv.org/abs/1906.05253}.

\bibitem[Zhu et~al.(2020)Zhu, Wong, Mandlekar, Mart\'{i}n-Mart\'{i}n, Joshi, Nasiriany, and Zhu]{robosuite2020}
Y.~Zhu, J.~Wong, A.~Mandlekar, R.~Mart\'{i}n-Mart\'{i}n, A.~Joshi, S.~Nasiriany, and Y.~Zhu.
\newblock robosuite: A modular simulation framework and benchmark for robot learning.
\newblock In \emph{arXiv preprint arXiv:2009.12293}, 2020.

\bibitem[Chang et~al.(2015)Chang, Funkhouser, Guibas, Hanrahan, Huang, Li, Savarese, Savva, Song, Su, Xiao, Yi, and Yu]{shapenet2015}
A.~X. Chang, T.~Funkhouser, L.~Guibas, P.~Hanrahan, Q.~Huang, Z.~Li, S.~Savarese, M.~Savva, S.~Song, H.~Su, J.~Xiao, L.~Yi, and F.~Yu.
\newblock {ShapeNet: An Information-Rich 3D Model Repository}.
\newblock Technical Report arXiv:1512.03012 [cs.GR], Stanford University --- Princeton University --- Toyota Technological Institute at Chicago, 2015.

\bibitem[Mandlekar et~al.(2021)Mandlekar, Xu, Wong, Nasiriany, Wang, Kulkarni, Fei-Fei, Savarese, Zhu, and Mart\'{i}n-Mart\'{i}n]{robomimic2021}
A.~Mandlekar, D.~Xu, J.~Wong, S.~Nasiriany, C.~Wang, R.~Kulkarni, L.~Fei-Fei, S.~Savarese, Y.~Zhu, and R.~Mart\'{i}n-Mart\'{i}n.
\newblock What matters in learning from offline human demonstrations for robot manipulation.
\newblock In \emph{arXiv preprint arXiv:2108.03298}, 2021.

\end{thebibliography}

\newpage 

% \appendices

\begin{appendices}
\section{Model Details}
\subsection{Data Modalities}
\label{appdx_data}
In simulation, we include an eye-in-hand $(224 \times 224 \times 3)$ and a third-person $(224 \times 224 \times 3)$ camera perspective. We also include low-dimensional proprioceptive data, including the robot's end-effector position, orientation, and gripper width. 

For the real robot experiments, we include an eye-in-hand $(256 \times 256 \times 3)$ image from a camera mounted to the wrist of the Franka-Emika robot. We also include proprioceptive data about the end-effector position, orientation, and gripper width. See Figure \ref{fig:value_demo} for an example of the camera angle on the real robot. 

In both real and simulated robot experiments, we use an absolute position/orientation action space, meaning that the policy outputs the target position and orientation. Empirically, we found that absolute position/orientation actions are easier to learn on the diffusion policy. 

\subsection{Diffusion Policy}
\label{diffusion_appdx}
We use the U-Net implementation of Diffusion Policy provided through Robomimic \cite{chiDiffusionPolicyVisuomotor2023a, robomimic2021}. The policy takes in images and feeds them through a Resnet-18 encoder (separate encoders for each camera) before concatenating the features with the low-dimensional proprioceptive state and encoding them through another MLP. The final feature is used to condition the noise model $\epsilon_\theta$. We sample action chunks of size 16 and execute 16 steps in an open-loop fashion before querying the diffusion policy for the next 16 steps. The action is in cartesian absolute space with one element of the action to control the robot gripper.

During training, we fit the prediction of $24$ action steps to the expert data. During testing, we sample the chunk of $24$ actions and execute $16$ in the environment before resampling. Following empirical results, we also chose to remove history from the Diffusion policy model. For most other hyperparameters, we used the default provided by the codebase. We use the same architecture for the real and simulated experiments. 

\textcolor{red}{}

\subsection{Value Function}
The Value Function is a visual encoder + MLP stack. It takes in the image(s), encodes them, and extracts a value distribution as a categorial distribution by feeding the image features through an MLP and softmax. The categorical distribution contains $50$ elements, each representing $2\%$ of progress. 

During training, we sample $s_t, t, T$ from the expert dataset. We can compute the relative progress as $t/T$, and we can use this to define the target distribution. We could round $t/T$ to the nearest $2\%$ and get bin $b$. Then, we could make the target distribution a one-hot vector at $b$. In practice, we found more success with a soft target. We find $b$ and fill $b-1, b, b+1$ with $1/3$ of the density each. This helps the value function output a broader distribution, which is helpful for our statistical methods.

\begin{figure*}[t]
    \centering %centers everything 
    \includegraphics[width = 0.9\textwidth]{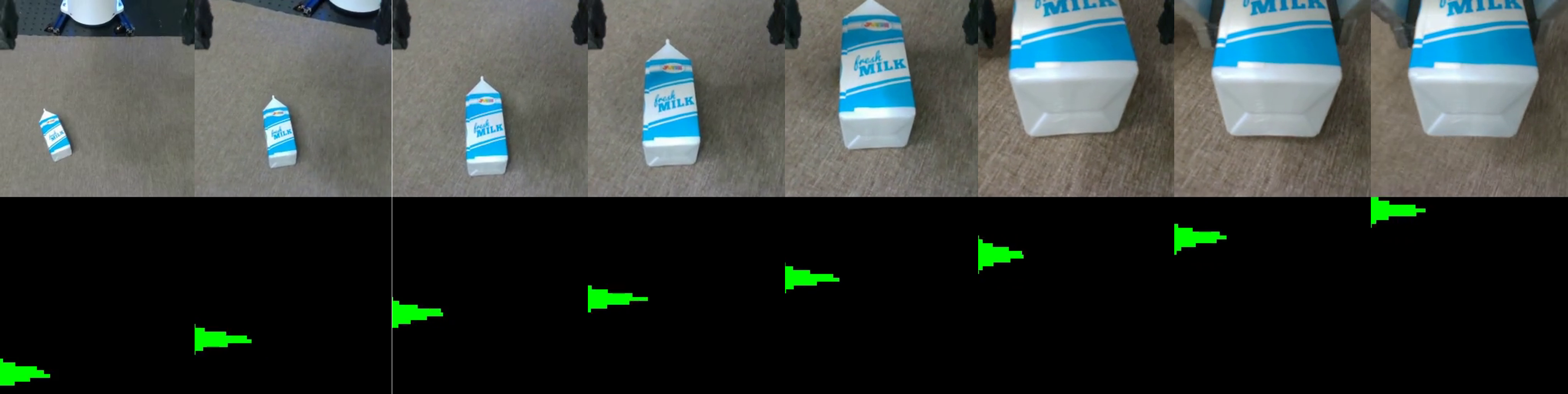}
    % \vspace{-0.2cm}
    \caption{\small \textbf{Demonstration of our trained value function on a successful trajectory.} In the distribution (second row), we visualize the outputs of the value function through a successful trajectory. The value function distribution increases as the robot gets closer to the milk jug. The value function learns a distance to task completion by images alone.}
        % \vspace{-0.5cm}
    \label{fig:value_demo} 
\end{figure*}

We do not feed proprioceptive data into the Value function to force it to leverage visual features (rather than simple end-effector locations) to measure progress. See Figure \ref{fig:value_demo} for a visualization of the trained value function behavior on a successful trajectory.

\begin{figure*}[t]
    \centering %centers everything 
    \includegraphics[width = 0.8\textwidth]{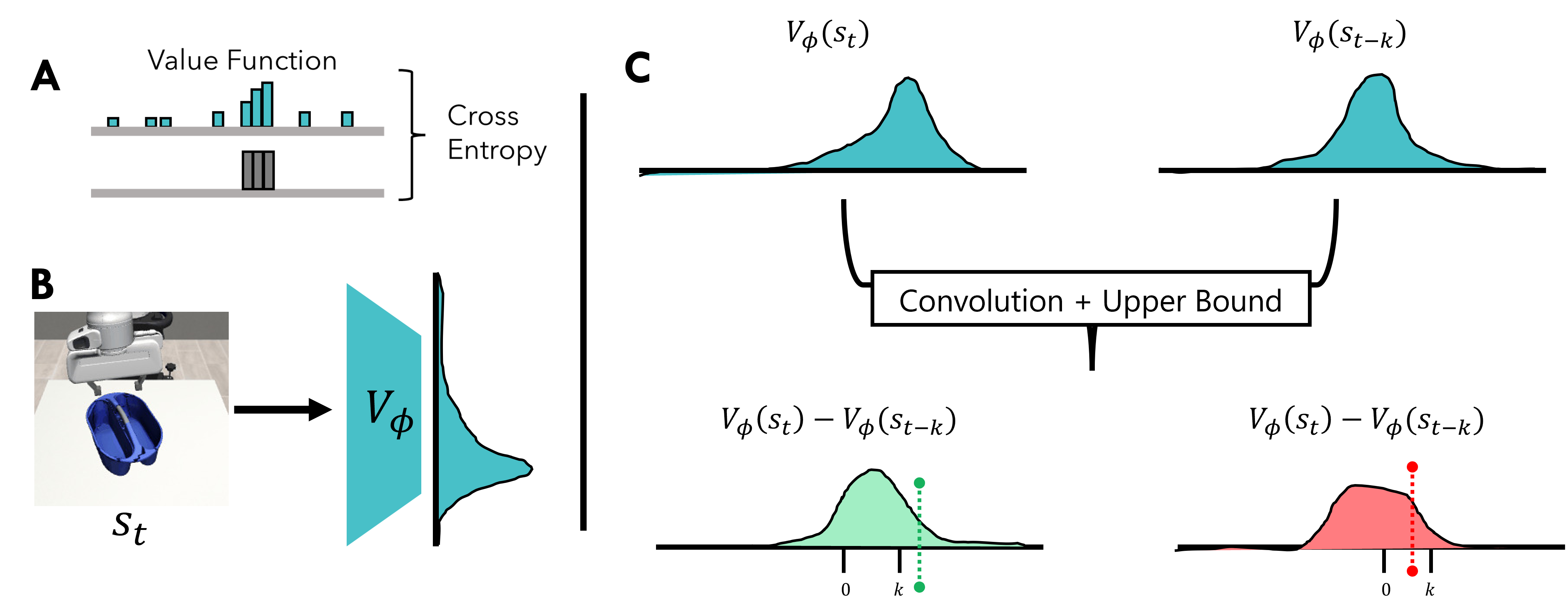}
    % \vspace{-0.2cm}
    
    \caption{\small \textbf{Details on training and using the value function.} We train $V_\phi$ to output a categorical distribution (B) by regressing the logit outputs to a softened one-hot vector (A). During test-time, we compute $V(s_t) - V(s_{t-k})$ by convolving the distributions and computing an upper bound on the difference (C). We show an example of acceptable progress (green, C) and not acceptable progress (red, C).}
        % \vspace{-0.5cm}
    \label{fig:value_method} 
\end{figure*}

\subsection{Tuning and Training}
We train both the diffusion policy and value function on the same datasets of expert demonstrations. We train the models until convergence, as evaluated on a set of validation trajectories. Concretely, we train diffusion policies for 200k gradient steps on the SimObjectLift task, 100k gradient steps on the CaddyGrasp task, and 400k on the RealObjectLift task. We train the value function for 400k gradient steps on the SimObjectLift task, 10k gradient steps on the CaddyLift task, and 120k steps on the RealObjectLift task. When evaluating on the whole system, we did a very light search on a set of value function checkpoints, and we picked the checkpoint that yielded the most stable results. In practice, the training of the diffusion policy and value functions are relatively stable and do not require anything more than a sanity check. 

\subsection{Implementing Progress Evaluation}

\begin{figure*}[htb]
    \centering %centers everything 
    \includegraphics[width = 0.95\textwidth]{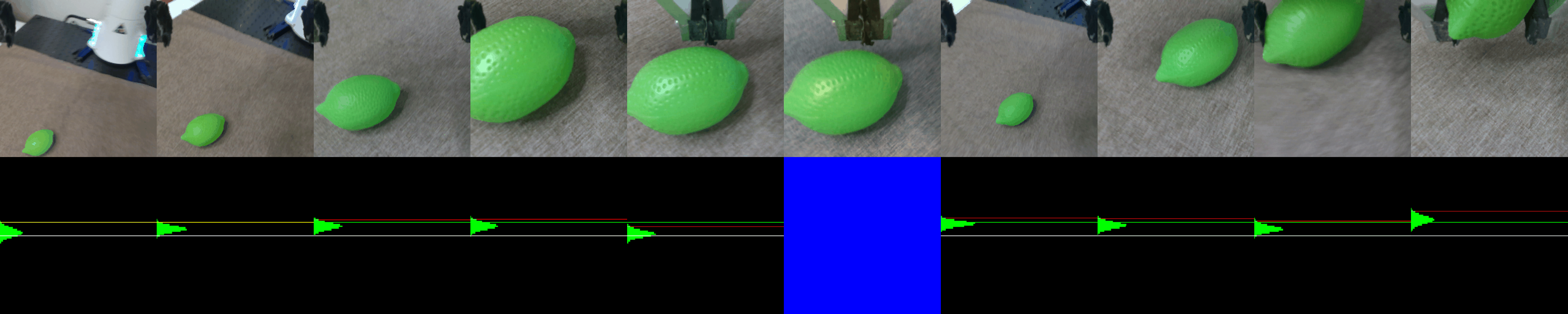}
    % \vspace{-0.2cm}
    \caption{\small \textbf{Demonstration of the progress evaluation on an initial failure.} In the distribution (second row), we visualize the delta distribution. The \textcolor{red}{red} horizontal line in the second row denotes the probabilistic upper bound, and the \textcolor{green}{green} horizontal line denotes the expected progress. The blue square indicates a recovery behavior.}
        % \vspace{-0.5cm}
    \label{fig:retrying_demo} 
\end{figure*}
We trained the value function to output a categorical distribution, which we can use to our advantage when we implement the progress evaluation part of our method. The expression $V(s_t) - V(s_{t-k})$ now becomes a convolution of two distributions. The new delta distribution $V(s_t) - V(s_{t-k})$ is a distribution of how the value function has changed in the last $t$ steps. A recovery and retrial should not be done unless we are quite sure that the robot has made a mistake, so we take a probabilistic upper bound of the delta distribution. We compare this upper confidence interval to the expected progress $k$. If the upper bound is less than $k$, then we can say with a high probability that the robot has made a mistake. See Figure \ref{fig:value_method} for a diagram of the distributional value function computations, and see Figure \ref{fig:retrying_demo} for a visualization of this progress evaluation on a real robot task. 

In practice, we derive the upper bound by computing the standard deviation of the delta distribution and adding two standard deviations to the mean of the delta distribution. 

We lightly tune the value of $k$, although we generally find success at $k = 20$. With our 20 Hz control setup, this corresponds to a one-second lookback time. Therefore, it is possible to find mistakes very quickly, down to a second of its occurrence.

\subsection{Implementing Skewing}
To skew the actions away from past mistake states, we keep track of these mistake states in the robot's proprioception space. Then, during execution, we sample $n$ different action chunks. For each action chunk, We compute the pairwise L2 distance between actions and avoidance points, and we find the minimum distance. This tells us how close the proposed action chunk takes us to the set of states to avoid. We pick the action chunk that maximizes this distance. Intuitively, the action chunk chosen from this algorithm is an action that avoids the past mistake states as much as possible while following the samples of the base policy. 

In simulation, we sample $10$ chunks and pick the best according to the above method. On a real robot, we sample $3$ times to reduce the time needed to query the diffusion policy.

\section{Environment Details}
\subsection{Generating / Collecting Data}
For the CaddyLift task, we collected $500$ demonstrations using human teleoperation through hand-held Oculus VR controllers. We used a random number generator to direct the demonstrator towards one part of the caddy to grasp, allowing us to approximate a uniform distribution of grasps on the caddy. 

For the SimObjectLift task, we collected $3600$ demonstrations by using a scripted policy that had access to privileged state information. The scripted policy attempt to pick up the object by finding an affordance and grasping it. We reject suboptimal demonstrations, which include slow grasps and trajectories that show grasp mistakes. The training set included grasps from 100 different objects sampled randomly from the Gazebo and Shapenet datasets.

For the RealObjectLift task, we use an internal dataset of object grasps on the Franka-Emika robot. We used a subset of the internal dataset that contains roughly $600$ demonstrations of grasps on a brown background. The training set included grasps from $>$10 different kitchen toy objects. 

In all tasks, we collect expert-only behavior in the datasets, which include very few demonstrations of mistake recovery. The scarcity of recovery demonstrations highlights the importance of \method in explicitly creating such capacities.  

\subsection{Simulation Evaluations}
\label{appdx_eval_details}
In CaddyLift, we introduced an invisible barrier during test-time that made two affordances invalid. The invisible barrier acts like another wall, preventing the grippers from reaching around the affordance. A robot that tries to grab a blocked affordance will find itself unable to maneuver its grippers around that affordance, but there are no visual deviations from the train data.

In SimObjectLift, we introduce ten novel objects sampled from the same training object distribution. These objects have never been seen in the dataset. For the adversarial task, we run a search over a set of novel objects and pick the objects that yielded the lowest success rates on the base policy.

For SimObjectLift and CaddyLift, we evaluated \method and baselines in simulation. With SimObjectLift, we used a horizon of 400 steps (not including recovery time) and a cap of 20 resets per episode, although nearly all episodes used less than 4 resets. For CaddyLift, we used a horizon of 600 steps (not including recovery time) and a cap of 20 resets per episode. In all simulation tasks, we provided all tested variants with the exact same set of starting objects and orientations. This matched pair comparison ensures that the differences in performance are purely explained by our method. To compute each success statistic, we collected 100 trials on three separate model seeds, allowing us to find the mean and sample variance of success rates. 

\subsection{Real Robot Setup}
We use a Franka-Emika Panda arm running on absolute positional control at 20 Hz. The images come from an Intel Realsense camera attached to the gripper using a custom 3D-printed mount. 

In the RealObjectLift-Train and RealObjectLift-Test, we evaluate the robot on a set of train objects and held-out test objects, respectively. These objects are placed randomly within a 10cm $\times$ 10cm table region, at random orientations. The test objects are similar kitchen toy objects but they have never been seen in the dataset.

To showcase our method's ability to improve the robustness of a base policy, we changed the dynamics in an adversarial task by wrapping a polyethylene sheet (originally from packing material) around the robot grippers and three difficult-to-grasp objects. This modification caused many existing grasping strategies to become less effective. For example, it is no longer optimal to grab the cheese by the wedge part, as the cheese can slip out as the grippers are closing. 

We strived to reduce the between-method variances as much as possible. Concretely, we evaluated all methods using the same set of start states. The object type, position, and orientation were made as similar as possible between trials across methods. To compute each success statistic, we collected 20 trials on the robot. 

% \color{red} % REMOVE THIS
\section{Additional Experiments and Discussion}
\label{addl_exp}

\subsection{How does \method Skew the Base Policy?}
\label{skewexp}

\begin{figure}[htb]
    \centering %centers everything 
    \includegraphics[width = 0.6\columnwidth]{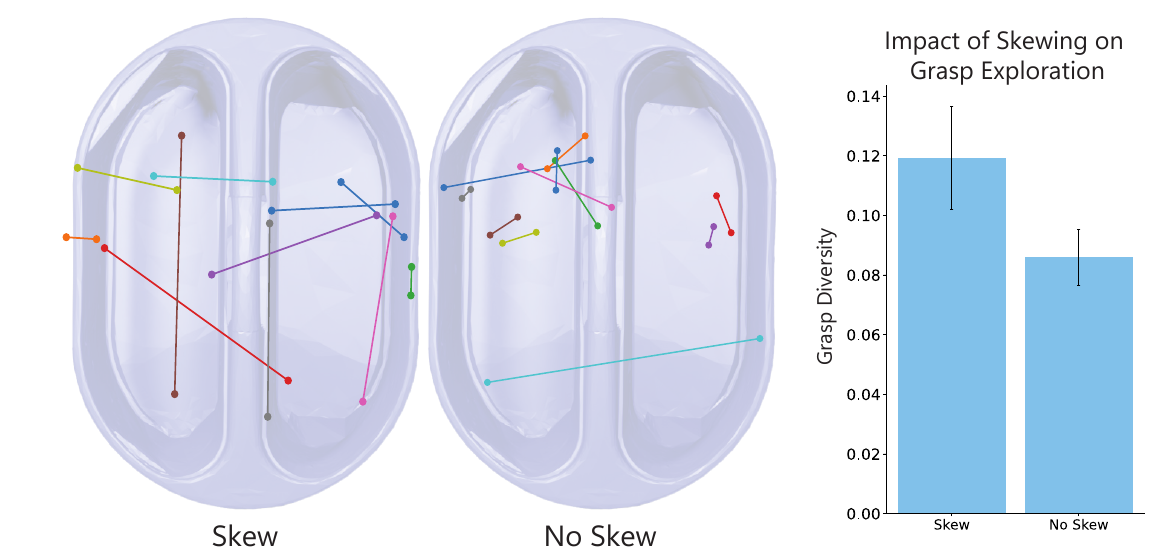}
    \caption{\small \textbf{Impact of Skewing on Strategy Diversity. } In this visualization, we look at the different grasp locations attempted by a robot with and without skewing from past mistakes (\textbf{left}). For a random sample of trajectories, we plot two consecutive grasp points (denoted by the pairs of points connected by a line segment). With skewing, the points are further apart, indicating greater diversity between attempts. These qualitative results are supported by a larger average distance between grasps (\textbf{right}).} 
    \label{fig:exp2} 
     \vspace{-0.2cm}
\end{figure}

For our second experiment, we are interested in looking at how we can improve the diversity of attempted strategies by skewing the policy away from past mistakes states (Section \ref{skew_method}). In Figure \ref{fig:exp2}, we plot a progression of grasp attempts in the \textbf{CaddyGrasp} environment. Consecutive grasps are often in similar locations of the caddy when we do not skew away from past (suboptimal) attempts. In contrast, the grasp locations for the skewed policy are further away and often span different handles.

We added the skewing component to our method because we wanted to give the policy the ability to adapt quickly to failures. Our suboptimal detection can give the robot multiple attempts to solve a scenario, but without some form of adaptation, the policy could potentially make the same mistake over and over. The qualitative and quantitative results shown in Figure \ref{fig:exp2} demonstrate the strategy diversity imparted by the skewing, which increases the likelihood of finding the correct strategy.

\subsection{Does \method Improve Timesteps to Success?}
\label{addl_exp_succ}

\begin{wraptable}{r}{0.65\textwidth}
% \begin{table}[htb]
    \vspace{-2em}
  \centering % \begin{center}
    \caption{\small \textbf{Comparing \method Timesteps to Success To Relevant Baselines.}}
     \resizebox{0.65\textwidth}{!}{
  \begin{tabular}{c|cccc} 
    \toprule     
      &  Base Policy  & Interval  & Ours & Ours \\ & No Recovery & Recovery & w/o Skew & (Full) \\
    \midrule
    \vspace{0.05cm}
    SimObjectLift Train & 171 $\pm$ 10 & 184 $\pm$ 12 & 160 $\pm$ 10 & \textbf{136 $\pm$ 7} \\
    SimObjectLift Test & 198 $\pm$ 9 & 202 $\pm$ 9 & \textbf{172 $\pm$ 4} & \textbf{172 $\pm$ 8} \\
    SimObjectLift Advr & 246 $\pm$17 & 232 $\pm$ 16 & 192 $\pm$ 20 & \textbf{177 $\pm$ 7} \\
    \midrule
    \textbf{Average} & 205 $\pm$ 12 & 206 $\pm$ 12 & 174 $\pm$ 11 & \textbf{162 $\pm$ 7} \\
    \bottomrule
  \end{tabular}
  \label{tab1}
  }
  \vspace{-1.5em}
\end{wraptable}

Success rate is an easy-to-calculate and directly-relevant metric. Therefore, in the main paper results, we chose to use success rate. There are also other possible metrics to compare \method, including the number of steps the robot takes until a success. If the robot fails, we assign the number of steps as the horizon maximum. See Table \ref{tab1} for results on the \textbf{SimObjectLift} task. Generally, our method reduces the number of steps needed until success. This is expected as our method prevents the robot from becoming stuck while also judiciously applying the recovery. The \textbf{Interval Recovery} tends to apply the recovery too frequently, which is reflected in the higher number of steps needed until success.

\subsection{More Details on SimObjectLift Experiment}
\label{simobjectlift_bonus}

\begin{wrapfigure}{r}{0.5\textwidth}
\vspace{-0.3cm}
    \centering %centers everything 
    \includegraphics[width = 0.45\textwidth]{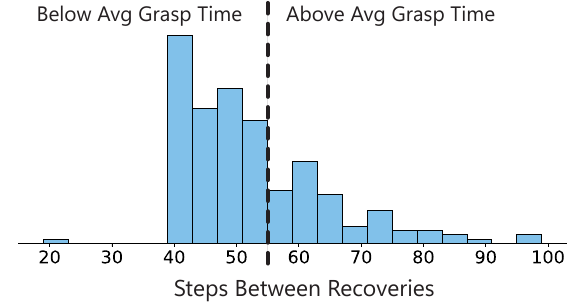}
    \caption{\small \textbf{\method Allows for Early Termination of Suboptimal Strategies. } Most recoveries of \method are tripped before the average time needed for an expert to grab and lift an object, showing that \method can detect mistakes quickly.} 
    \label{fig:histogram} 
     \vspace{-0.5cm}
\end{wrapfigure}
The base policy may try to make a grasp, fail, and then hover above the object with its grippers closed. In contrast, our method generally triggers a recovery immediately when it becomes apparent that an object has not been grasped successfully (e.g. grippers are closed with no object between them). This gives the robot many more attempts to grasp the object. We also observe many instances of early recovery (Figure \ref{fig:histogram}), where our algorithm finds suboptimality in the trajectory before the robot attempts lift with a bad grasp, reducing the time per trial and the risk of irrecoverably changing the environment (e.g., knocking the object off the table). We detail all simulated results in Figure \ref{fig:exp3}.

\subsection{More Details on RealObjectLift Experiment}
\label{realobjectlift_bonus}
 A common failure mode is a misgrasp, but another failure mode is an incorrect reach. The robot operates on wrist camera only, which means that an initial mistake during the reaching process could move the entire object out of frame. In these situations, the base policy fails by reaching toward a part of the table that does not contain the object. This mistake is quite out of distribution, and the robot rarely recovers. In contrast, the strategy evaluation component of our method is quick to detect when the object has left the frame and triggers a recovery that brings the object back into the frame for another attempt. Similar to the simulation results, this early termination allows our method to outperform an Interval Recovery baseline by stopping suboptimal attempts much faster. 

In the \textbf{test} and \textbf{adversarial} scenarios, we evaluate the robot on objects that are more difficult to grasp, like a smooth breadstick or a wedge of cheese. These objects introduce a new failure mode: object slippage during grasping. We exacerbate this failure mode in our \textbf{adversarial} variant when we reduce the friction of the robot grippers and objects with plastic film (Figure \ref{fig:envs}). When faced with these dynamic failures, our strategy evaluation component is quick to trigger a recovery---often within a second of an observed slippage (Figure \ref{fig:exp1}). The \textbf{adversarial} variant also demonstrates the importance of skewing. With very different dynamics, a failure often happens because a previously feasible strategy is now suboptimal (e.g., trying to pick up a slippery carrot by the tapered end). Skewing the policy away from past mistakes helps our method improve over the interval recovery baseline. We detail all real robot results in Figure \ref{fig:exp3}. 

\subsection{More Details on DoorOpening Experiment}
\label{addl_exp_robot}
To demonstrate \method, the many of the main experiments used grasping tasks. However, \method is not limited to such tasks. To demonstrate the versatility of \method, we conducted an additional real robot experiment on an articulated \textbf{DoorOpening} task that requires the robot to rotate a handle and then pull the handle outwards to open the door (Figure \ref{fig:envs}). We collected expert demonstrations that showed a variety of successful strategies. The handle was rotated using different contact strategies and the unlocked handle was also grasped in different locations. We used these demonstrations to train the Value function and the base policy. We evaluated the robot on the same setup, and as can be seen in Figure \ref{fig:exp3}, \method can improve performance by more than double the baseline success rate. Unfortunately, due to unforeseen circumstances, the real-world environment was lost before we could conduct our full sweep of experiments, but the existing two results show promising evidence that \method can boost performance of complicated, long-horizon task policies.

% \begin{table}[tb]
%   \centering % \begin{center}
%     \caption{\small \textbf{Results of applying \method to an articulated door task.}}
%     \begin{tabular}{c|cc} 
%     \toprule     
%       &  Base Policy  & Ours  \\
%     \midrule
%     \vspace{0.05cm}
%     DoorOpening & 6 / 20 & 15 / 20 \\
%     \bottomrule
%   \end{tabular}
%   \label{tab2}
%   \vspace{-0.3cm}
% \end{table}

Qualitatively, the base robot policy frequently became stuck after trying to rotate the door handle by contacting the handle too close to the pivot, which required excessive force. Our method could detect the slowdown of progress and perform a recovery (pulling the arm back). When the arm came back to the handle, the robot often switched strategies and used a different contact that allowed for greater leverage on the handle. Through its mistake-recovery-retry sequence, \method is able to get the robot unstuck. 

\subsection{More Details on Ablations}
\label{ablation}
We need to disentangle the role of \textit{strategy evaluation} from the role of \textit{recoveries}. We make this comparison by looking at \textbf{Ours w/o Skew} with \textbf{Interval Recovery} in Figures \ref{fig:exp3}. The interval recovery baseline triggers a recovery after a certain number of timesteps. We lightly tune these timesteps by using the average number of steps of an expert trajectory and adding some buffer steps to account for slightly slower progress. Critically, this tuning of the baseline must be done for every policy and environment, while \method is automatic. Under a fixed horizon time limit, our method outperforms the interval reset baseline in all but one variant. We hypothesize that our method has a benefit because it can detect mistakes quickly. Indeed, as seen in Figure \ref{fig:histogram}, a distribution of strategy trial lengths shows many trials that were terminated before the robot attempted to lift an object. From these results, we conclude that a strategy evaluation-based recovery can benefit success rates if we operate under a fixed horizon. 

We discussed the qualitative role of skewing in Section \ref{skewexp}. By comparing between \textbf{Ours w/o Skew} and \textbf{Ours (Full)} in Figures \ref{fig:exp3}, we also see quantitative differences. These differences are most apparent on difficult scenarios like the \textbf{adversarial} object variants. We hypothesize that this trend is due to the increased need for diverse strategy exploration when the test-time scenarios differ more from the training scenarios.

From the three pairwise comparisons described above, we can conclude that both components of our method are important to the final performance improvement, with benefits being more pronounced in harder scenarios. 
Such a trend indicates that our method improves the robustness of the base policy.

\end{appendices}

% \color{black}

\end{document}